\documentclass[11pt, logo, onecolumn, copyright, colorlinks=true, allcolors=blue]{nvidiatechreport}
\usepackage{graphicx} 
\usepackage{subcaption,ragged2e}
\usepackage[round]{natbib}

\usepackage[normalem]{ulem}
\usepackage[utf8]{inputenc} 
\usepackage[T1]{fontenc}    
\usepackage{url}
\usepackage{tikz}
\usepackage{enumitem}
\usetikzlibrary{positioning}
\usetikzlibrary{calc}
\usepackage{array}
\usepackage{booktabs}
\usepackage{wrapfig}
\usepackage{caption} 
\usepackage{listings}
\usepackage{adjustbox}
\usepackage{footnote}
\usepackage{multirow}
\usepackage{amsmath}
\usepackage{amsfonts}
\usepackage{breakcites}
\usepackage{blindtext}
\usepackage{svg}
\usepackage[most]{tcolorbox}
\usepackage{tabularx}
\usepackage{graphicx} 
\usepackage{xcolor}
\usepackage{colortbl}
\newcommand{\newparagraph}[1]{\noindent\textbf{#1\hspace{0.5em}}}

\newcommand{\ourmodelfull}{NVIDIA Nemotron 3 Nano\xspace}
\newcommand{\ourmodel}{Nemotron 3 Nano\xspace}
\newcommand{\ourbasemodel}{Nemotron 3 Nano 30B-A3B Base\xspace}
\newcommand{\ourgenrm}{Qwen-3-Nemotron-235B-A22B-GenRM}
\newcommand{\ourfinalmodel}{Nemotron 3 Nano 30B-A3B\xspace}

\title{\ourmodel: Open, Efficient Mixture-of-Experts Hybrid Mamba-Transformer Model for Agentic Reasoning}
\author{\large NVIDIA}
\date{}

\begin{document}

\begin{abstract}
\large \textbf{Abstract.}
\normalsize
We present \ourfinalmodel, a Mixture-of-Experts hybrid Mamba-Transformer language model. \ourmodel was pretrained on 25 trillion text tokens, including more than 3 trillion new unique tokens over Nemotron 2, followed by supervised fine tuning and large-scale RL on diverse environments. \ourmodel achieves better accuracy than our previous generation Nemotron 2 Nano while activating less than half of the parameters per forward pass. It achieves up to 3.3$\times$ higher inference throughput than similarly-sized open models like GPT-OSS-20B and Qwen3-30B-A3B-Thinking-2507, while also being more accurate on popular benchmarks. \ourmodel demonstrates enhanced agentic, reasoning, and chat abilities and supports context lengths up to 1M tokens. We release both our pretrained \ourbasemodel and post-trained \ourfinalmodel checkpoints on Hugging Face.
\end{abstract}

\maketitle

\section{Introduction}
\label{sec:intro}

\begin{figure}[ht]
    \centering
    \includegraphics[width=\linewidth]{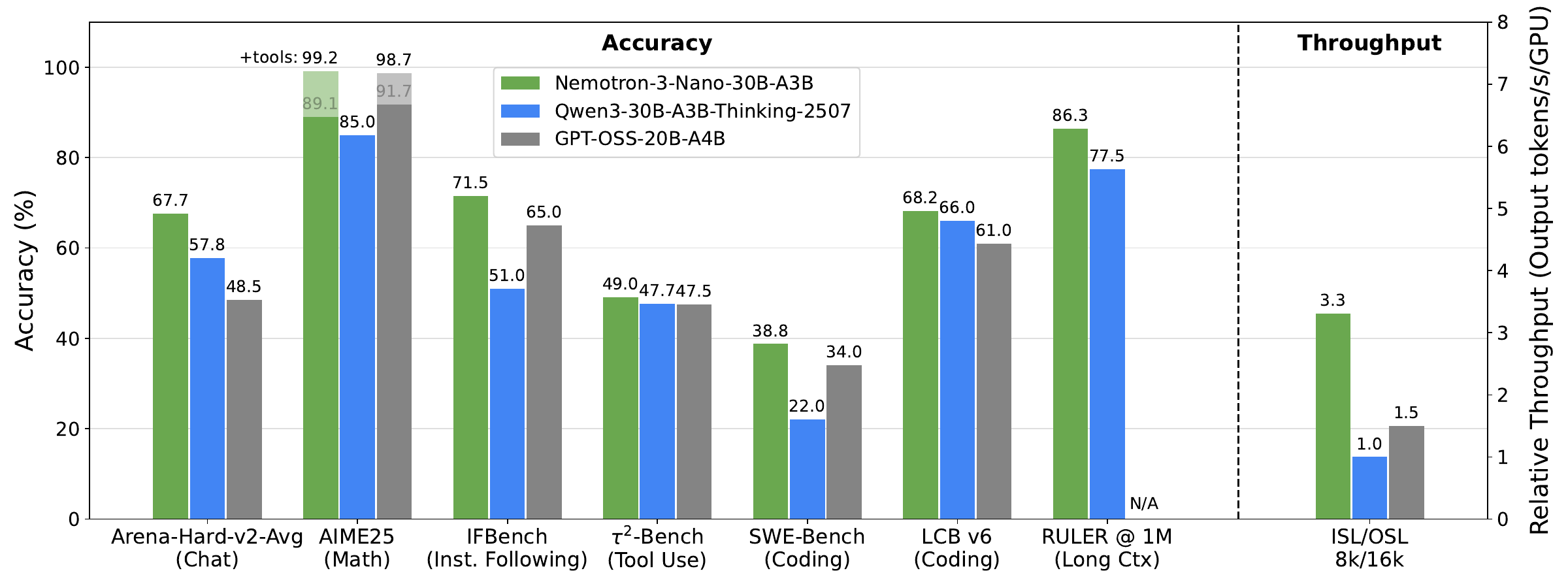}
    \caption{Accuracy and throughput comparisons of \ourmodel with Qwen3-30B-A3B-Thinking-2507 and GPT-OSS-20B. \ourmodel achieves on-par or better accuracies across multiple benchmarks. RULER scores for 1M context length are available only for \ourmodel and Qwen3 since GPT-OSS-20B has a context length of 128K tokens. Further, on 8K input / 16K output setting, \ourmodel provides inference throughput that is 3.3$\times$ higher than Qwen3-30B-A3B-Thinking-2507 and 2.2$\times$ higher than GPT-OSS-20B. We measured throughput on a single H200 GPU with vLLM and TRT-LLM and used the best out of the two for each model. We used the OpenHands harness to evaluate SWE-Bench.}
    \label{fig:intro}
\end{figure}

We present \ourmodelfull, a Mixture-of-Experts (MoE) hybrid Mamba-Transformer model~\citep{lieber2024jambahybridtransformermambalanguage} with agentic, reasoning, and chat capabilities. Like previous generations~\citep{nvidia2025nemotronhfamilyaccurateefficient,nemotronnanov2}, \ourmodel uses a combination of Mamba-2~\citep{dao2024transformersssmsgeneralizedmodels} and Grouped-Query-Attention (GQA)~\citep{ainslie2023gqatraininggeneralizedmultiquery}. In addition, \ourmodel uses Mixture-of-Experts~\citep{shazeer2017outrageously} layers to scale model parameters sparsely and achieve significant improvements on the inference-throughput-to-accuracy frontier. We use a granular MoE architecture~\citep{dai2024deepseekmoe} with a learnt MLP router that activates 6 out of 128 experts~(\S\ref{sec:arch}). \ourmodel totals 31.6B parameters out of which only 3.2B are activated per forward pass (3.6B including embeddings). \ourmodel achieves better or on-par accuracy compared to GPT-OSS-20B~\citep{openai2025gptoss120bgptoss20bmodel} and Qwen3-30B-A3B-Thinking-2507~\citep{yang2025qwen3technicalreport} as shown in Figure~\ref{fig:intro}. Further, on the 8K input / 16K output token scenario,
\ourmodel provides 2.2$\times$ and 3.3$\times$ faster inference throughput compared to GPT-OSS-20B and Qwen3-30B-A3B-Thinking-2507 respectively. \ourmodel also supports context lengths up to 1M tokens, outperforming both GPT-OSS-20B and Qwen3-30B-A3B-Instruct-2507 on RULER across different context lengths. Along with the model weights, we provide the recipe, code, and most of the data we used to train the model.

We pretrained our base model, \ourbasemodel, using the Warmup-Stable-Decay~\citep{hu2024minicpm} learning rate schedule on 25 trillion tokens of text data spanning 15 categories (\S\ref{sec:pretrain_data}). We divided pre-training into 2 phases with 23.5 trillion tokens of diverse data in the first phase, followed by 1.5 trillion tokens of high-quality data in the second phase (\S\ref{section:blend}). Our base model achieves better accuracy than equivalent-sized Qwen3-30B-A3B-Base on most academic benchmarks across Code, Math, Long Context, General Knowledge, and Commonsense Understanding categories. We do not compare the accuracy of our base model to GPT-OSS-20B because no base model was released with it.  Our model also achieves significantly better inference throughput than Qwen3-30B-A3B~(3.3$\times$) and GPT-OSS-20B~(2.2$\times$) on generation heavy 8K input / 16k output scenario when tested on a single H200 GPU. We measured throughput using the best configuration available for H200 GPUs with both vLLM and TRT-LLM and used the better of the two for each model. We used \texttt{FP8} for both weights and activations for throughput measurement of \ourmodel and Qwen3. We used \texttt{mxfp4} for weights and \texttt{bfloat16} for activations for GPT-OSS-20B.

We post-trained \ourmodel using three approaches: supervised fine tuning (SFT) (\S\ref{subsec:SFT}), multi-environment reinforcement learning from verifiable rewards (RLVR) (\S\ref{subsec:RLVR}), and reinforcement learning from human feedback (RLHF) (\S\ref{subsec:RLHF}). During SFT, we trained \ourmodel on a diverse set of chat, agentic, and reasoning traces to imbue the model with reasoning budget control, reasoning on/off control, and tool-integrated reasoning capabilities. During RLVR, we trained on all environments simultaneously, resulting in a smooth and uniform improvement in model capabilities. During RLHF, we utilized a large and accurate generative reward model (GenRM) to enhance the performance of \ourmodel on key chat benchmarks. 

We also quantized \ourmodel from \texttt{bfloat16} to \texttt{FP8} using post training quantization (PTQ). This helps achieve higher inference throughput with minimal loss in accuracy~(\S\ref{subsec:quantaccuracy}).

Along with this report, we are releasing the model recipes\footnote{\href{https://github.com/NVIDIA-NeMo/Nemotron}{https://github.com/NVIDIA-NeMo/Nemotron}} and publishing the following:

\newparagraph{Checkpoints}
\begin{itemize}
    \item \href{https://huggingface.co/nvidia/NVIDIA-Nemotron-3-Nano-30B-A3B-FP8}{\texttt{\ourfinalmodel FP8} \includegraphics[height=0.9em]{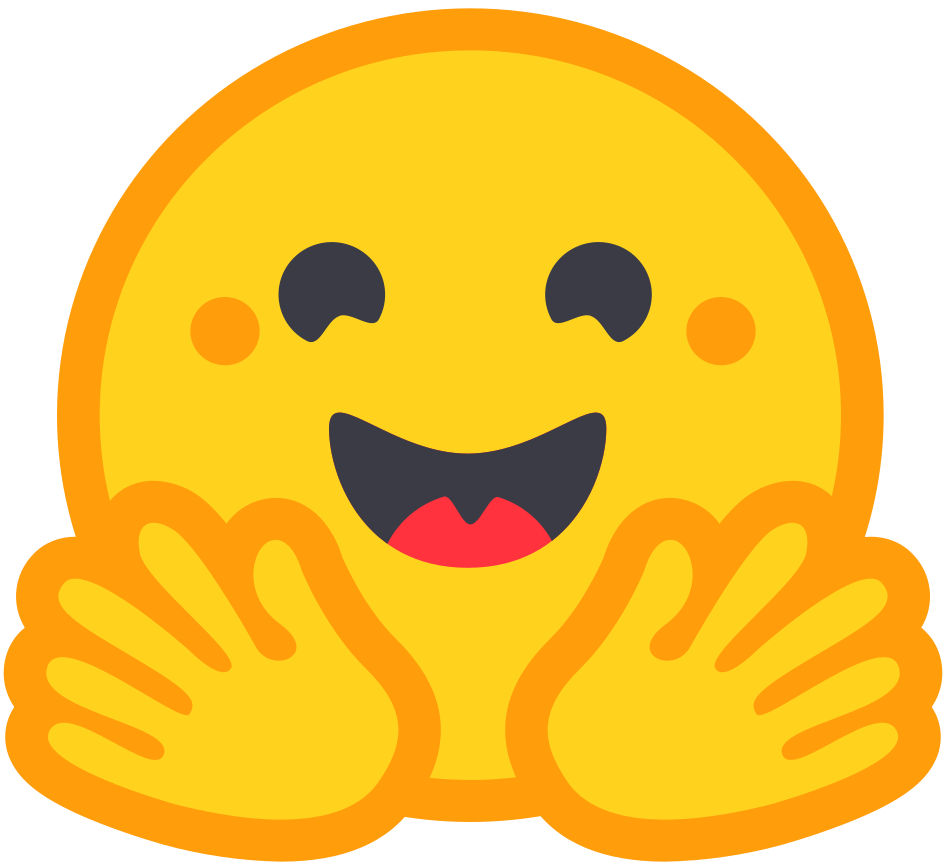}} : the final post-trained and FP8 quantized model 
    \item \href{https://huggingface.co/nvidia/NVIDIA-Nemotron-3-Nano-30B-A3B-BF16}{\texttt{\ourfinalmodel BF16} \includegraphics[height=0.9em]{assets/huggingface-color.png}} : the post-trained model
    \item \href{https://huggingface.co/nvidia/NVIDIA-Nemotron-3-Nano-30B-A3B-Base-BF16}{\texttt{\ourbasemodel BF16} \includegraphics[height=0.9em]{assets/huggingface-color.png}} : the pre-trained base model
    \item \href{https://huggingface.co/nvidia/Qwen3-Nemotron-235B-A22B-GenRM}{\texttt{\ourgenrm} \includegraphics[height=0.9em]{assets/huggingface-color.png}} : the GenRM used for RLHF
\end{itemize}

\newparagraph{Data}
\begin{itemize}
    \item \href{https://huggingface.co/datasets/nvidia/Nemotron-CC-v2.1}{\texttt{Nemotron-CC-v2.1} \includegraphics[height=0.9em]{assets/huggingface-color.png}} : 2.5 trillion new English tokens from Common Crawl, including curated data from 3 recent snapshots, synthetic rephrasing, and translation to English from other languages.
    \item \href{https://huggingface.co/datasets/nvidia/Nemotron-CC-Code-v1}{\texttt{Nemotron-CC-Code-v1} \includegraphics[height=0.9em]{assets/huggingface-color.png}} : A pretraining dataset consisting of 428 billion high-quality code tokens obtained from processing Common Crawl Code pages using the Lynx + LLM pipeline from \href{https://huggingface.co/datasets/nvidia/Nemotron-CC-Math-v1}{\texttt{Nemotron-CC-Math-v1} }. Preserves equations and code, standardizes math equations to LaTeX, and removes noise.
    \item \href{https://huggingface.co/datasets/nvidia/Nemotron-Pretraining-Code-v2}{\texttt{Nemotron-Pretraining-Code-v2} \includegraphics[height=0.9em]{assets/huggingface-color.png}} : Refresh of curated GitHub code references with multi-stage filtering, deduplication, and quality filters. Large-scale synthetic code data.
    \item \href{https://huggingface.co/datasets/nvidia/Nemotron-Pretraining-Specialized-v1}{\texttt{Nemotron-Pretraining-Specialized-v1} \includegraphics[height=0.9em]{assets/huggingface-color.png}} : Collection of synthetic datasets for specialized areas like STEM reasoning and scientific coding.
    \item \href{https://huggingface.co/collections/nvidia/nemotron-post-training-v3}{\texttt{Nemotron-SFT-Data} \includegraphics[height=0.9em]{assets/huggingface-color.png}} : Collection of new \ourmodel SFT datasets.
    \item \href{https://huggingface.co/collections/nvidia/nemo-gym}{\texttt{Nemotron-RL-Data} \includegraphics[height=0.9em]{assets/huggingface-color.png}} : Collection of new \ourmodel RL datasets.

\end{itemize}



We divide the remainder of the report into 3 sections: Pre-training~(\S\ref{sec:pretraining}), Post-Training~(\S\ref{sec:alignment}), and Quantization~(\S\ref{sec:quantization}). 

\section{Pretraining}
\label{sec:pretraining}

In this section, we highlight the key features of \ourbasemodel, including its architecture, hyperparameters, and the data used for pretraining. We also show that \ourbasemodel achieves better accuracy than other public state-of-the-art models across a suite of benchmarks.

\subsection{Model Architecture}
\label{sec:arch}

\begin{figure}[!t]
\centering
\includegraphics[width=0.9\linewidth]{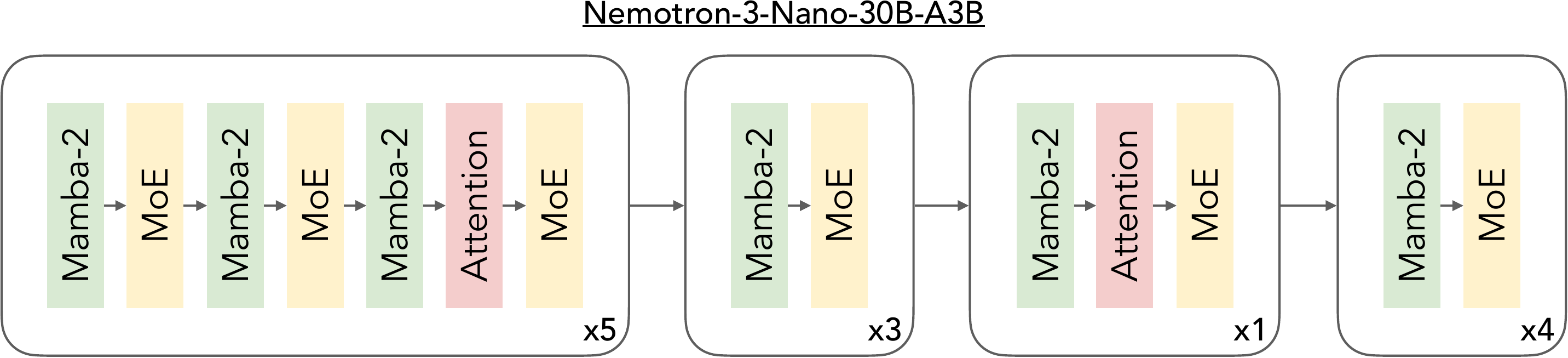}
\caption{\ourmodel layer pattern. We use a hybrid Mamba-Transformer architecture similar to the previous generation of Nemotron models. In addition, we scale the model sparsely by using MoE layers instead of standard FFN layers.}
\label{fig:layer-pattern}
\end{figure}

\begin{table*}[!hbt]\small\centering
\renewcommand{\arraystretch}{1.2} 
\setlength{\tabcolsep}{6pt} 
\begin{tabular}{l|c}
\toprule
\textbf{Model} & \textbf{\ourbasemodel} \\
\midrule
Num Layers         & 52  \\
Model Dimension             & 2688 \\
\midrule
Q-heads             & 32 \\
KV-heads             & 2 \\
Head Dimension             & 128 \\
\midrule
Mamba State Dimension             & 128 \\
Mamba Groups             & 8 \\
Mamba Heads         & 64 \\
Mamba Head Dimension      & 64 \\
\midrule
Expert Dimension           & 1856 \\
Total Routable Experts             & 128 \\
Number of Activated Experts             & 6 \\
Number of Shared Experts             & 2 \\

\bottomrule
\end{tabular}
\caption{\ourmodel Architecture}
\label{table:model_arch}
\end{table*}

\ourbasemodel builds upon the hybrid Mamba-Transformer architecture of our older Nemotron-H~\citep{nvidia2025nemotronhfamilyaccurateefficient} and Nemotron 2 Nano~\citep{nemotronnanov2} models by replacing the standard FFN layers with sparse Mixture-of-Experts (MoE)~\citep{shazeer2017outrageously} layers. The MoE layers help us achieve better accuracy at a fraction of the active parameter count. \ourbasemodel contains 31.6B total parameters out of which 3.2B are active (3.6B including embeddings) per forward pass. To achieve the best accuracy, we use a granular MoE architecture along with shared experts~\citep{dai2024deepseekmoe}. For the MoE layers, we use squared ReLU activation and a standard learnt MLP router with sigmoid gating. We do not use any positional embeddings, dropout, or bias on linear layers. We use RMSNorm for normalization and un-tie embedding and projection weights.




Table~\ref{table:model_arch} and Figure~\ref{fig:layer-pattern} show the key architectural details of \ourmodel.
\subsection{Pretraining Data}
\label{sec:pretrain_data}

In this sub-section, we describe new datasets that we added to our pretraining corpus since Nemotron Nano 2. We are releasing the vast majority of the new data on HuggingFace,
divided into four main datasets. We describe each of these in more detail below.
%

\subsubsection{Nemotron-CC-Code-v1}

We first filtered out the code pages in Common Crawl based on a fast pattern matching code classifier for webpages. We then constructed our high-quality code pretraining corpus by applying a modified version of the Nemotron-CC-Math pipeline~\citep{karimi2025nemotronccmath} to Common Crawl pages containing code. 

Starting from raw HTML, we rendered each document using Lynx, which reliably preserved code layout, indentation, and inline technical elements. The resulting text was processed by an LLM-based cleaning stage using the Phi-4 model, which removed boilerplate while strictly retaining code snippets, configuration blocks, API references, and mathematical expressions. To ensure that only programming-relevant documents are included, we applied a lightweight code-quality relevance classifier, filtering out non-technical pages and retaining documents with substantial or complete code content. This pipeline produced a 427.92B-token corpus in which equations are standardized to LaTeX, code blocks are preserved with structural fidelity, and noise is minimized. Compared to previous extraction approaches that often corrupt or truncate code examples, our method reliably recovered complete code snippets and technical context at scale. 

\subsubsection{Nemotron-Pretraining-Code-v2}
We sourced additional code from GitHub for repositories we identified as missing from our existing corpus in addition to collecting recent data with a cut-off date of April 15, 2025. We used the same pipeline as described in ~\cite{nemotronnanov2} to curate the data and we remove exact and near-duplicate files already present in our existing corpus.

In addition to our raw source-code corpus, we synthetically generated additional mixed natural-language and source code documents using the Qwen3 32B LLM. Similar to our approach described in ~\cite{nvidia2025nemotronhfamilyaccurateefficient}, we prompted the model to generate question and answer pairs using our new source-code data as seeds. Additionally, we prompted the model to generate student-teacher (Python only) and code-review (Python/C++) style dialogue grounded with a combination of code snippets and full source files.

Following the code-rewriting work presented in ~\cite{fujii2025rewriting}, we also found that using LLMs to rephrase source code improved downstream code-generation accuracies. Using Qwen3 32B, we rephrased all of our raw Python source code using a combination of the Style-Guided Code Rewriting (SGCR) and Self-Contained Optimization Rewriting (SCOR) prompts ~\citep{fujii2025rewriting}, as well as our own prompt with similar intent. To ensure high-quality LLM rephrasing, as a post-processing step, we checked for syntax errors and assessed code-quality improvements using the Pylint Python linter for each of the rewritten files.

While LLM-based source-code rewriting can be observed as a transformation of the original source-code to an improved version, we extended this concept and applied it to source-code files from one language to another (i.e., code transpilation). Using Qwen3 32B we found that C++ tokens produced from Python using this transpilation procedure improved downstream C++ code-generation accuracy and thus served as a useful augmentation to our C++ subset. We applied this Python to C++ transpilation procedure to all Python source files in our source-code corpus.

\subsubsection{Nemotron-CC-v2.1}
For general English web crawl data, we added three more recent Common Crawl snapshots on top of \href{Nemotron-CC-v2}{https://huggingface.co/datasets/nvidia/Nemotron-CC-v2} (CC-MAIN-2025-18, CC-MAIN-2025-21, CC-MAIN-2025-26), prepared with the same Nemotron-CC recipe~\citep{su2024nemotroncctransformingcommoncrawl}. For all of the synthetic rephrasing, we used Qwen3-30B-A3B~\citep{yang2025qwen3technicalreport}. Just as for Nemotron Nano 2, we trained only on the Medium-Quality, Medium-High-Quality, and High-Quality buckets.

Previously, we rephrased only the High-Quality subset of Common Crawl data.
To further expand our corpus of unique high-quality tokens, we applied five prompts~\citep{su2024nemotroncctransformingcommoncrawl} to the Medium-High-Quality data from 110 Common Crawl snapshots (CC-MAIN-2013-20 - CC-MAIN-2025-26), resulting in 2.1T new tokens.

Finally, we employed a new strategy to source high-quality English tokens by translating to English from other languages using Qwen3-30B-A3B. 
We first translated documents from the latest three Common Crawl snapshots available at that time (CC-MAIN-2024-51, CC-MAIN-2025-08, and CC-MAIN-2025-18) in 9 languages (Chinese, French, German, Italian, Japanese, Polish, Portuguese, Russian, Spanish) to English.
After that, we applied the Nemotron-CC ensemble of quality classifiers to retain only High-Quality and Medium-High-Quality documents from this translated subset.
Additionally, we applied four of the five Nemotron-CC rephrasing prompts to the high-quality data to generate more unique tokens.
After training of \ourbasemodel was already underway, we found that some uninformative translated documents (e.g., daily conversations, ads) were receiving high scores from the Nemotron-CC quality classifiers.
To address this, for the released version of this dataset, we performed one additional pass of LLM-based quality filtering that removed approximately 10.6~\% of tokens, which slightly improved accuracies across benchmarks in an internal ablation.

Overall, we curated or generated over 2.5T new tokens from Common Crawl data.

\subsubsection{Nemotron-Pretraining-Specialized-v1}

This dataset comprises various synthetic datasets that are specialized for specific topics like STEM Reasoning or scientific coding. We describe the subsets in more detail below.

\newparagraph{Synthetic Wikipedia Data} We revised English Wikipedia articles using Qwen3-30B-A3B-Instruct-2507 to improve clarity and formatting. We discarded disambiguation and redirect pages and removed References, See also, Notes, and External Links sections. We also instructed the model to remove any irrelevant content such as uncleaned HTML elements.

\newparagraph{Synthetic Math Textbook Data}
We generated well-structured educational textbook-style sections from Nemotron-CC-Math~\citep{karimi2025nemotronccmath}.
We evaluated the mathematical content in each document and classify it into an educational level (e.g., grade school, middle school, high school) based on multiple factors such as involved mathematical concepts and complexity.
We kept documents containing mathematical content at the undergraduate level and above and developed each into a textbook-style section with diverse educational features such as definitions and illustrative examples.

\newparagraph{Synthetic Scientific Coding Data} Using STEM-related documents retrieved from Nemotron-CC as the seed data, we synthesized two types of documents:
(1) Code-embedded article: A comprehensive, in-depth, and well-formatted article that explores and implements a non-trivial, graduate- or research-level scientific or mathematical algorithm in Python;
(2) Computational coding problem: An advanced, computational, graduate- or research-level coding problem with Python solution. The main problem is decomposed into 5 to 15 logically ordered non-trivial substeps, each solved by an individual function. We extract the main problem, dependencies, substep descriptions, and each function's signature, docstring, body, and return statement and exclude examples where any of these components are missing.

\newparagraph{Synthetic Cross-Domain Code Data}
To generate more diverse and complex code data, we develop a novel approach we call \textit{InfiniByte} that cross-breeds multiple datasets together. When applied to code, InfiniByte creates entirely new programming problems by bringing together concepts from different fields to pose never before seen questions. In doing so, InfiniByte fills the problem space between disparate domains, generates questions at the boundary of model capabilities, and mimics how science is often advanced at the intersection of two or more fields.

Starting with a curated list of competitive coding problems from our groundbreaking OpenCodeReasoning dataset (\cite{ahmad2025opencodereasoning}), we systematically inject concepts from datasets across mathematics (OpenMathReasoning, \cite{moshkov2025aimo2}), physics (Physics Big, \cite{physics_big}), chemistry (IChO, \cite{icho_ipho}), and other sciences. We generate multiple problem candidates per (problem, concept) combination, select the best problem candidate, based on LLM-as-critic rubric that tests for clarity, difficulty, and adherence to the employed cross-breeding strategy. We then generate solutions to each new coding problem using a reasoning model such as \texttt{Qwen3-235B-A22B-Thinking-2507} \citep{yang2025qwen3technicalreport}. We cross-breed with two different strategies in mind:
\begin{enumerate}
    \item[1.] Obfuscate without really changing the original problem (this is common in competitive coding problems and other competitions).
    \item[2.] Complicate by actually making the new problem much more complex: the resulting problem is more challenging as it requires reasoning across multiple concepts to solve it. 
\end{enumerate}

The InfiniByte data generation pipeline was implemented in NeMo Data Designer (\cite{nemo-data-designer}), NVIDIA’s state-of-the-art synthetic data generation framework. This allowed our complex pipeline to benefit from the compound AI approach of the framework in order to enforce proper concept grounding via Jinja templating, guarantee structured outputs required at all stages, incorporate feedback loops, as well as perform data validation and automated retries. 

\newparagraph{Synthetic STEM Reasoning}
To reinforce complex reasoning capabilities within STEM domains, we built the Reasoning Question-Answer (RQA) dataset. Our goal in the creation of RQA was two-fold: 
\begin{enumerate}
    \item[i)] Demonstrate advanced scientific reasoning and instruction following that can be further reinforced in post-training, as shown in \cite{akter2025frontload}.
    \item[ii)] Reinforce correlations between advanced topics that are otherwise rarely observed in web-scale data. 
\end{enumerate}
The dataset was generated in four steps. First, we targeted diverse and advanced scientific texts as seed data. Starting from the STEM subset of the Essential-Web web-scale dataset \citep{essentialweb}, we filtered the dataset using the Essential-Web taxonomy to documents that met the following criteria:
\begin{itemize}
    \item Undergraduate or graduate education level.
    \item No extraction artifacts, no missing content.
    \item Advanced reasoning depth.
    \item High or exceptional technical correctness.
    \item Leverages one of the Bloom cognitive processes: \emph{Analyze}, \emph{Evaluate} or \emph{Create}.
    \item Leverages one of the Bloom knowledge domains: \emph{Conceptual}, \emph{Procedural} or \emph{Metacognitive}.
    \item In the English language and over 1000 characters.
\end{itemize}

This filtering resulted in approximately 14 million documents. Next, we used hierarchically stratified sampling on document topics to trade-off between seed document volume and diversity. Leveraging the Free Decimal Correspondence (FDC) numerical topic code from the Essential-Web taxonomy, documents were ordered in hierarchical round-robin fashion across multiple orders of magnitude in the FDC code, from high-level topic domains (e.g. Physics, Chemistry, Math, Computer Science) to lower-level subdomains (e.g. Thermodynamics, Quantum Mechanics). Using this approach, we could apply any cutoff N to the seed documents to ensure maximum diversity for a given volume of documents; while we generated RQA samples for the first 9 million samples, we ultimately chose to use the first 4.5 million for training. To limit the length of each seed document, we post-processed documents over 4096 characters in length to extract a random contiguous text chunk consisting of <4096 characters.

Each seed document was presented as context to \texttt{Qwen3-235B-A22B-Thinking-2507}, which was prompted to use the STEM content as inspiration for a difficult (yet answerable) graduate-level scientific reasoning question. The model was instructed to ensure that the question did not require access to the original seed passage to answer. Examples were discarded if they failed to produce a question within 8192 reasoning tokens.

Finally, this question was presented to \texttt{Qwen3-235B-A22B-Thinking-2507} to answer in a second generation step, without including the seed passage as context. The resulting reasoning trace and answer were filtered to remove model-specific idiosyncrasies, limited to 8192 characters, and concatenated with the question to produce a single RQA example. The two-step generation process was designed to maximally engage the teacher model's reasoning capabilities, both in generating a difficult question from the seed document and in answering its own question. The resulting pretraining dataset consists of 4.3 million RQA demonstrations for a total of approximately 31.7 billion unique tokens. 

To make further use of the stratified STEM seed documents, we also produced a diverse QA (DQA) version of the dataset, using the first 9 million seed documents in stratification order for a total of approximately 8 billion tokens. The STEM DQA dataset was built by using the DQA prompt \& generation procedure as demonstrated in Nemotron-CC \citep{su2024nemotroncctransformingcommoncrawl}, which concatenates a contiguous text chunk from the source document with short-form question-answer pairs. We utilized \texttt{Qwen3-30B-A3B} to generate these QA pairs.

Both RQA and DQA data generation pipelines were implemented in NeMo Data Designer (\cite{nemo-data-designer}).

\newparagraph{SFT-style data.}
We included new and refreshed SFT datasets in pretraining for code, math, and STEM, just as for Nemotron Nano 2. Detailed synthesis methods and pipelines can be found in prior works~\citep{toshniwal2024openmathinstruct, moshkov2025aimo2,bercovich2025llamanemotronefficientreasoningmodels,ahmad2025opencodereasoning,ahmad2025opencodeinstruct,majumdar2024genetic}.
We also incorporated a set of additional math and code SFT samples from AceReason-Nemotron-1.1~\citep{liu2025acereason}. This collection encompasses a wide range of prompt sources, including NuminaMath~\citep{numina_math_datasets}, OrcaMathWordProblems~\citep{mitra2024orca}, MathInstruct~\citep{yue2023mammoth}, and MetaMathQA~\citep{yu2023metamath} for math tasks, as well as TACO~\citep{li2023taco}, APPs~\citep{hendrycksapps2021}, OpenCoder-Stage2~\citep{Huang2024OpenCoderTO}, and OpenCodeReasoning~\citep{ahmad2025opencodereasoning} for coding tasks. The responses for these prompts are generated by DeepSeek-R1~\citep{deepseekai2025deepseekr1incentivizingreasoningcapability}.

\subsection{Data Mixture and Ordering}
\label{section:blend}

Our pretraining corpus spans fifteen data categories.
The largest component is web crawl data, which we subdivide into five quality-based groups following the Nemotron-CC taxonomy~\citep{su2024nemotroncctransformingcommoncrawl}: crawl-medium, crawl-medium-high, syn-crawl-medium-high, crawl-high, and syn-crawl-high, representing medium, medium-high, high,  crawl data.
Beyond web crawl, the mixture also includes math, Wikipedia, code, nemotron-cc-code, academic text, Crawl$++$, multilingual data, and synthetic SFT-style datasets, the latter further grouped into general-sft, stem-sft, and code-sft categories.
Crawl$++$ comprises the OpenWebText, BigScience and Reddit datasets. 
Our multilingual data has nineteen languages:
Arabic, Chinese, Czech, Danish, Dutch, Finnish, French, German, Hebrew, Hindi, Italian, Japanese, Korean, Portuguese, Polish, Russian, Spanish, Swedish, and Thai. 
We design our data mixtures to balance coverage and quality by assigning comparable weight to sources of similar estimated quality. Higher-quality datasets are prioritized accordingly, receiving greater weight in the blend. Additional details on our dataset quality assessment and mixture construction methodology can be found in \cite{feng2024maximizedataspotentialenhancing} and \cite{nvidia2025nemotronhfamilyaccurateefficient}.

\begin{figure}[htbp]
    \centering
    \begin{subfigure}{0.49\textwidth}
        \centering
        \includegraphics[width=\linewidth]{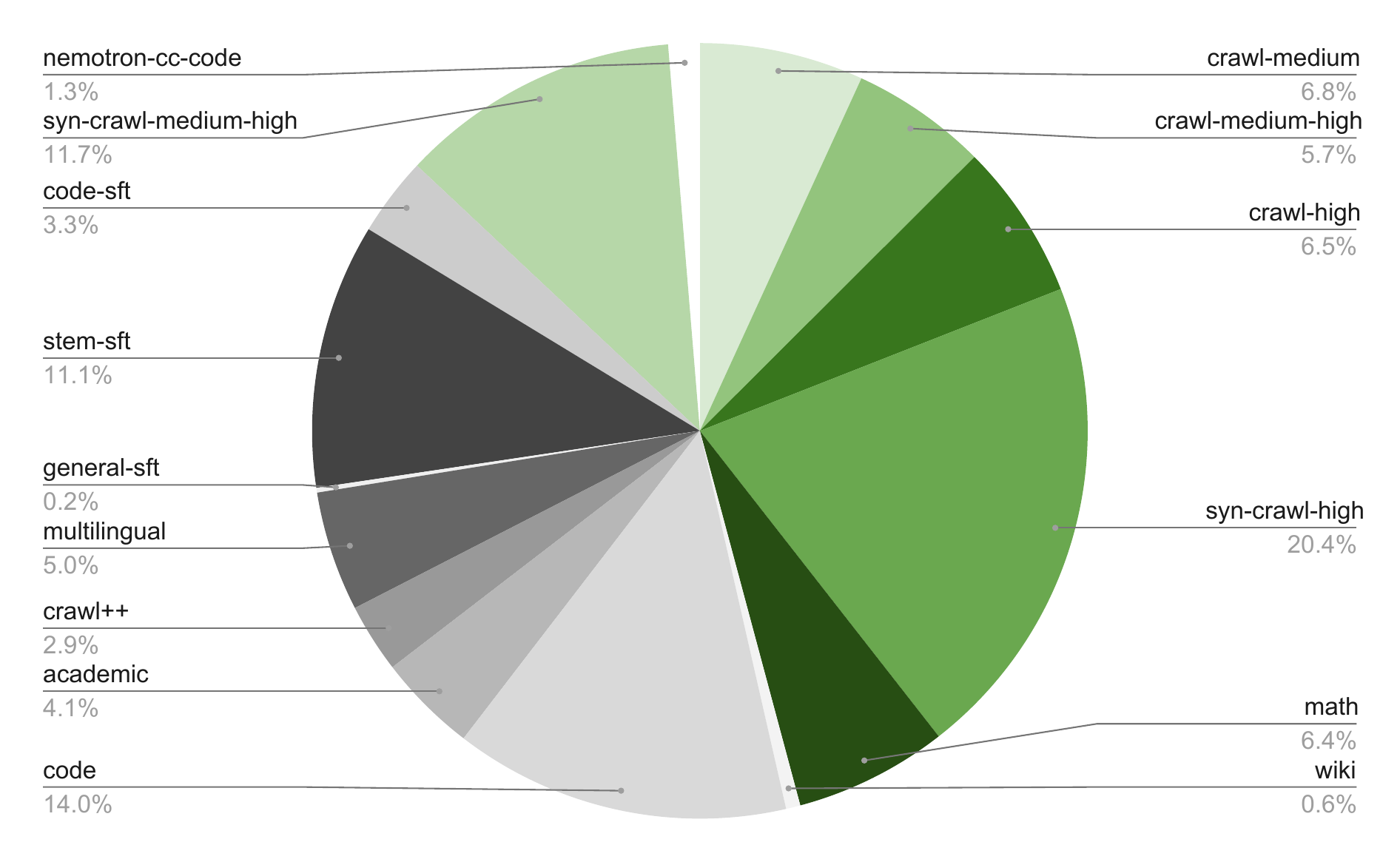}
        \caption{Data mixture of Phase 1.}
        \label{fig:phase1-blend}
    \end{subfigure}
    \begin{subfigure}{0.49\textwidth}
        \centering
        \includegraphics[width=\linewidth]{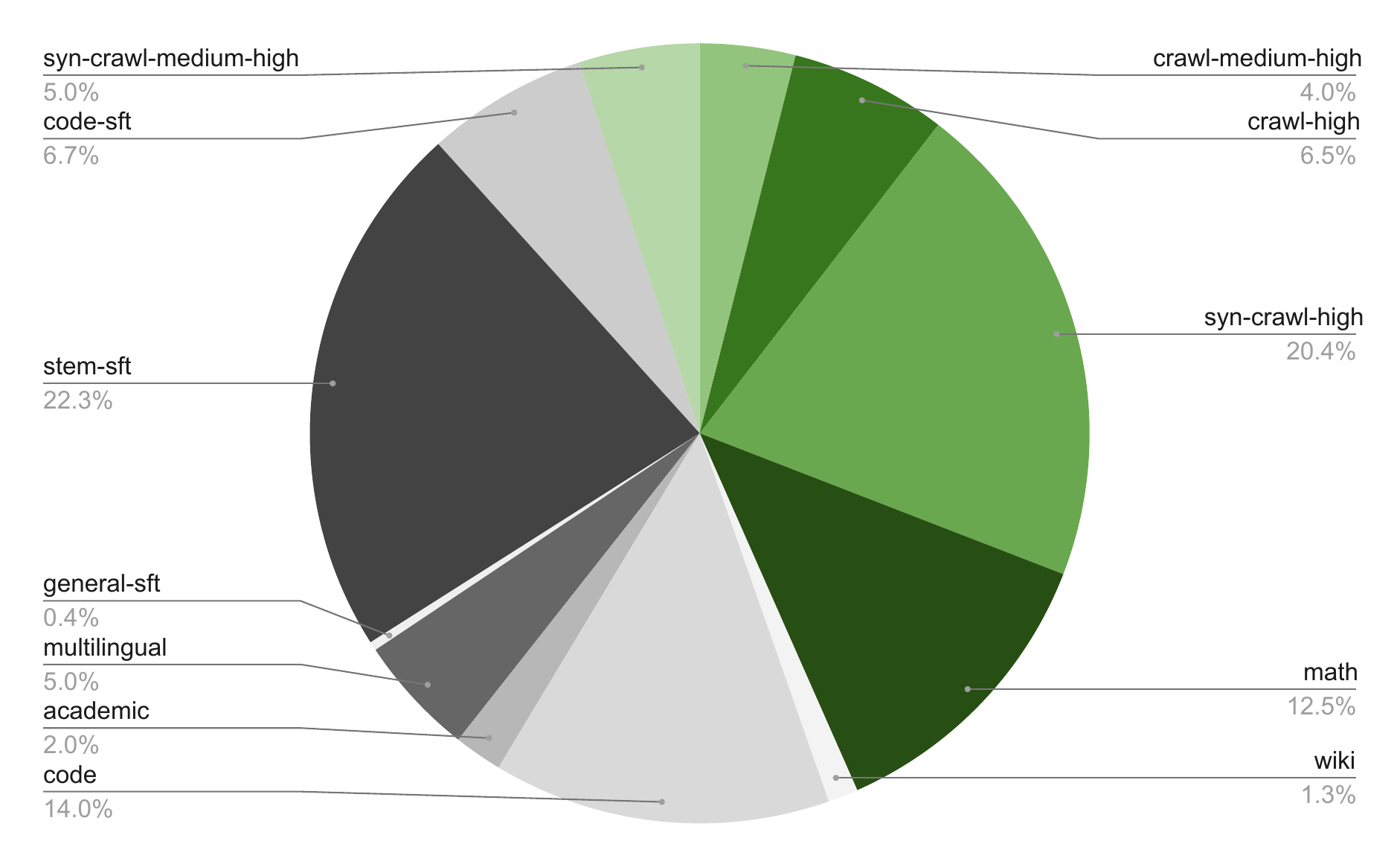}
        \caption{Data mixture of Phase 2.}
        \label{fig:phase2-blend}
    \end{subfigure}
    \caption{Data mixtures for each phase of pre-training.}
    \label{fig:phase-blends}
\end{figure}

We used a curriculum based on two phases to pre-train \ourbasemodel. 
In the first phase, we used a data mixture that promotes diversity in data; in the second phase, we primarily use high-quality datasets (e.g., Wikipedia). 
We switched to the second phase at the 94\% point of training. 
The data mixtures used in each phase are shown in Figure~\ref{fig:phase-blends}. 

\subsection{Hyperparameters}\label{sec:pre-train-hyperparams}
We pretrained \ourbasemodel using the Warmup-Stable-Decay learning rate (LR) schedule for a total of 25 trillion tokens. We warmed up the LR over $8.4$ billion tokens to a maximum of $10^{-3}$. We maintained the maximum LR for 80\% of training (20 trillion tokens) and then finally decayed to a minimum of $10^{-5}$ during the last 20\% of training (5 trillion tokens). We used the AdamW~\citep{loshchilov2017decoupled} optimizer with weight decay of $0.1$, $\beta_1=0.9$, and $\beta_2=0.95$. We pretrained the model with a sequence length of 8192 and a batch size of 3072, resulting in roughly 25 million tokens per batch. For the MoE layers, we used DeepSeek's aux-loss-free load balancing strategy~\citep{wang2024auxiliary,deepseekai2025deepseekv3technicalreport} with an update rate of $10^{-3}$ in conjunction with the standard load balancing loss~\citep{lepikhin2020gshard}. We used a load balancing loss coefficient of $10^{-4}$.

\subsection{Long-Context Extension}\label{sec:pre-train-long-context}

Similar to Nemotron 2 Nano, we added a long-context phase (LC-Phase) at the end of pretraining. In the LC-Phase, we performed continuous pretraining (CPT) to equip the base model with long-context ability. We used a constant learning rate of $10^{-5}$ and global batch size of 48. We used 8-way context parallelism, 8-way tensor parallelism, 8-way expert parallelism, and 4-way pipeline parallelism to train on H100 GPUs. We reused the long-context document QA dataset from Nemotron Nano 2, but scaled it to make it 3$\times$ larger. We also added a small amount of synthetic retrieval-focused data to the CPT data blend, with a maximum sequence length of 256k tokens, to help improve subset of RULER style tasks. We allocated the document QA and synthetic retrieval-focused data to 20\% and 1\% in the Phase LC data blend, with the remaining 79\% being downscaled Phase 2 data. We initially tried performing CPT on data batches with only sequence lengths of 524,288 (512k) tokens, but found that short-context benchmark scores were impacted to a small extent. Consequently, we used a mixture of 512k and 4k sequences, which resulted in improved short-context benchmark scores, especially MMLU-Pro and Code, while also improving long-context benchmark scores. LC-Phase used a total of 121 billion tokens.

\subsection{Base Model Evaluations}
\label{subsec:base_model_evals}

\begin{table}
\centering
\small
\setlength{\tabcolsep}{7pt}
\renewcommand{\arraystretch}{1.15}

\begin{tabular}{l|cc}
\toprule
\textbf{Task} &
\multicolumn{1}{c}{\textbf{Qwen3-30B}} &
\multicolumn{1}{c}{\textbf{N-3-Nano}} \\
&
\multicolumn{1}{c}{\textbf{A3B-Base}} &
\multicolumn{1}{c}{\textbf{30B-A3B Base}} \\
\midrule

\rowcolor{black!5}
\multicolumn{3}{l}{\textbf{General Knowledge}} \\
MMLU (5-shot, acc) & \textbf{81.07} & 78.56 \\
MMLU-Pro (5-shot, CoT EM) & 61.71 & \textbf{65.05} \\
AGIEval-En (3/5-shot, CoT acc) & 63.12 & \textbf{68.32} \\

\midrule
\rowcolor{black!5}
\multicolumn{3}{l}{\textbf{Code}} \\
HumanEval (0-shot) & 70.73 & \textbf{78.05} \\
MBPP-Sanitized (3-shot) & 73.15 & \textbf{75.49} \\
\midrule
\rowcolor{black!5}
\multicolumn{3}{l}{\textbf{Math}}  \\
GSM8K (8-shot, acc) & 89.01 & \textbf{92.34} \\
MATH (4-shot, acc) & 61.14 & \textbf{82.88} \\
MATH-500 (4-shot, avg@32) & 55.08 & \textbf{78.63} \\
\midrule
\rowcolor{black!5}
\multicolumn{3}{l}{\textbf{Commonsense Understanding}} \\
ARC-Challenge (25-shot, acc\_norm) & \textbf{94.45} & 91.89 \\
HellaSwag (10-shot, acc\_norm) & 83.14 & \textbf{85.56} \\
OpenBookQA (0-shot, acc\_norm) & 44.80 & \textbf{46.20} \\
PIQA (0-shot, acc\_norm) & 81.01 & \textbf{84.33} \\
WinoGrande (5-shot, acc) & 78.22 & \textbf{79.64} \\
\midrule
\rowcolor{black!5}
\multicolumn{3}{l}{\textbf{Reading Comprehension}} \\
RACE (0-shot, acc) & \textbf{90.05} & 88.04 \\
\midrule
\rowcolor{black!5}
\multicolumn{3}{l}{\textbf{Multilingual}}  \\
MMLU Global Lite (5-shot, avg acc) & \textbf{76.84} & 74.47 \\
MGSM (8-shot, avg acc) & 82.53 & \textbf{83.00} \\
\midrule
\rowcolor{black!5}
\multicolumn{3}{l}{\textbf{Long Context}}  \\
RULER (64K, 0-shot, acc) & 63.55 & \textbf{87.50} \\
RULER (128K, 0-shot, acc) & 60.69 & \textbf{82.92} \\
RULER (256K, 0-shot, acc) & - & \textbf{75.44} \\
\bottomrule
\end{tabular}
\caption[Comparison of Qwen3 vs N-Nano-3.1]{
    Comparison of \textbf{Qwen3-30B-A3B-Base} and \textbf{\ourbasemodel}. Best results are marked in bold.
}
\label{tab:model-comparison}
\end{table}

Table \ref{tab:model-comparison} presents a comprehensive accuracy comparison across general knowledge, code, math, commonsense understanding, reading comprehension, multilingual, and long context benchmarks. Evaluation settings adhered to standard community protocols to ensure fair comparison. All evaluation results were collected via Nemo Evaluator SDK\footnote{\url{https://github.com/NVIDIA-NeMo/Evaluator}} and LM Evaluation Harness\footnote{\url{https://github.com/EleutherAI/lm-evaluation-harness}}. For reproducibility purposes, more details on the evaluation settings can be found in the Nemo Evaluator SDK configs folder\footnote{\url{https://github.com/NVIDIA-NeMo/Evaluator}}, and the open source container on LM Evaluation Harness packaged via NVIDIA's Nemo Evaluator SDK used for evaluations can be found here\footnote{\url{https://catalog.ngc.nvidia.com/orgs/nvidia/teams/eval-factory/containers/lm-evaluation-harness}}.

For the MATH-500 task, we employed a sampling strategy to report the \texttt{avg@32} score (\texttt{pass@1} estimated from 32 samples). For the rest of the tasks, we report accuracy (\texttt{acc}) or normalized accuracy (\texttt{acc\_norm}) obtained via greedy decoding (\texttt{temperature = 0}). For code evaluations, HumanEval and MBPP, we apply the same sanitization method as in Evalplus\footnote{\url{https://github.com/evalplus/evalplus}}. Few-shot settings varied by benchmark, ranging from 0-shot for HumanEval to 25-shot for ARC-Challenge. Multilingual capabilities were evaluated on MMLU Global Lite (averaging across German, Spanish, French, Italian, Japanese, Korean, Portuguese, and Chinese) and MGSM (averaging across German, Spanish, French, Japanese, Russian, and Chinese).

To gain deeper insights into the model’s capabilities, we further evaluate the model on two variants of MMLU-redux (See Appendix~\ref{appendix:mmlu_redux_evaluation}).

\section{Post-Training}
\label{sec:alignment}
In comparison to Nemotron Nano 2, we significantly scale up the compute in post-training for \ourmodel. 
Noticeably \ourmodel is our first effort to scale up reinforcement learning (RL) in the post-training stage. 
This RL scale up is empowered by multi-environment reinforcement learning (discussed in Sections \ref{subsec:SFT} to \ref{subsec:RLHF}), where we train on all environments simultaneously for the first time. We adopted Nemo-Gym, a RL training environment orchestration framework with a large collection of RL environments; this is integrated with Nemo-RL as the RL training framework as discussed in Section \ref{subsubsection:RLinfra} \citep{nemo-gym, nemo-rl}. We open source Nemo-Gym and Nemo-RL to enable the broader community to facilitate large-scale RL training, as well as collaborative and distributed RL environment building.

In the rest of this section, we discuss the post-training methodology for \ourmodel, which includes supervised finetuning (SFT) in \S\ref{subsec:SFT}, multi-environment reinforcement learning in \S\ref{subsec:RLVR}, and reinforcement learning from human feedback (RLHF) in \S\ref{subsec:RLHF}. The final evaluation results can be found in \S\ref{subsec:final_model_evals}. Our post-training methodology results in best-in-class performance in a variety of reasoning and agentic tasks, along with token efficiency, reasoning on/off control, reasoning budget control, and tool-integrated reasoning capabilities.  

\subsection{Supervised Fine Tuning}
\label{subsec:SFT}
Since the release of Nemotron 2 Nano, we have significantly improved our SFT strategy. We increased dataset quality and diversity, adding a wide variety of new data with an emphasis on multi-step and multi-turn agentic tasks. Different from the SFT data in the pre-training stage, the SFT stage data is more focused on agentic tasks and has the chat-template applied. We release the majority of our training data and open source our SFT codebase.

\subsubsection{Chat Template}
We allow using \ourmodel in reasoning or non-reasoning mode through the chat template. 

In reasoning mode, we alter the reasoning flow for the following conversation scenarios:
\begin{itemize}
    \item \textit{Multi-Step}: In a series of assistant model calls, the existing reasoning tokens are preserved to allow the model to re-use existing reasoning for subsequent step.
    \item \textit{Multi-Turn}: When a user message is introduced, any reasoning from previous turns are dropped.
\end{itemize}

For tool calling, we use XML-style special tags to reduce character escaping, following the observations of GLM-4.5 \citep{5team2025glm45agenticreasoningcoding} and Qwen3-Coder \citep{yang2025qwen3technicalreport}.


\begin{figure}[t]
    \centering
    \includegraphics[width=10cm]{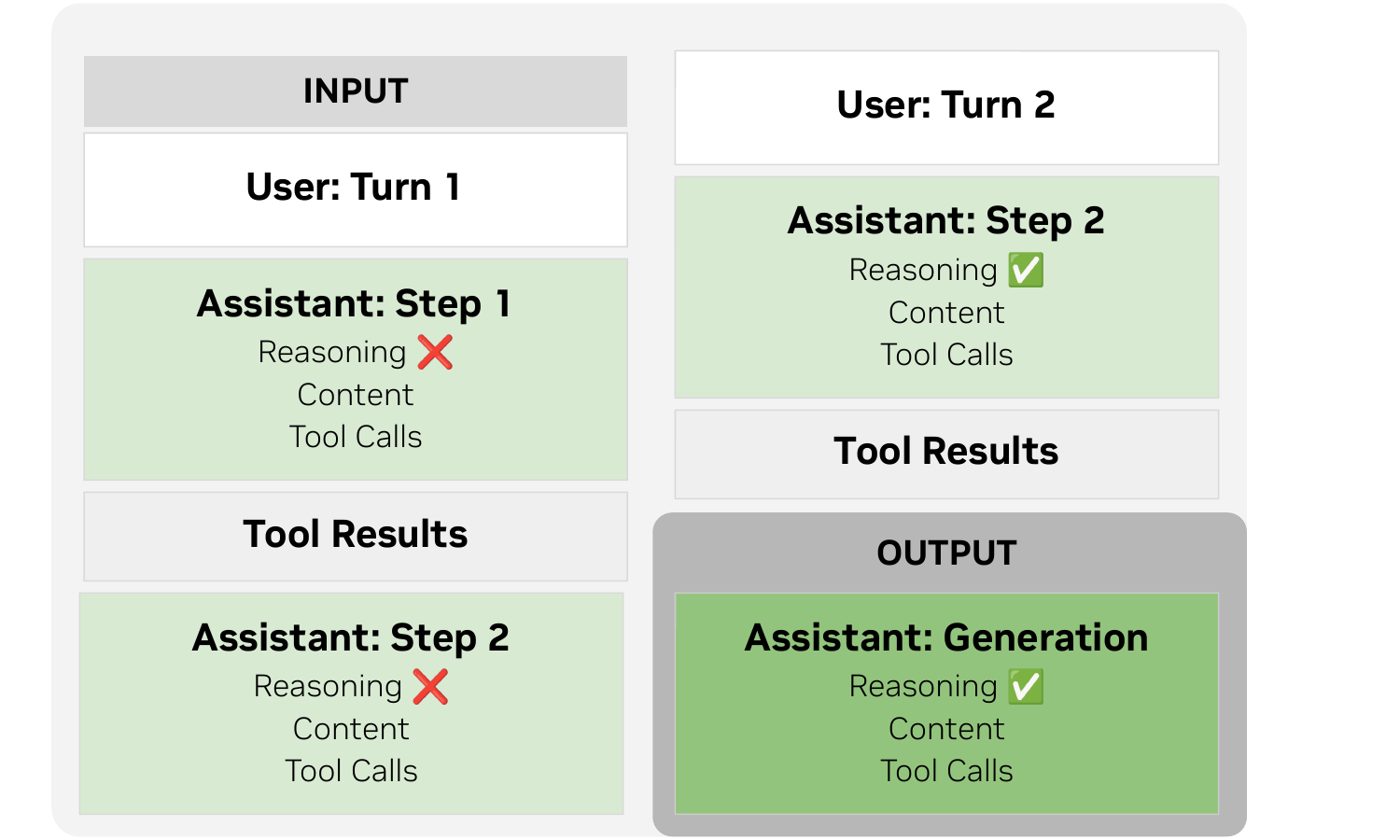}
    \caption{Example prompt materialization using the \ourmodel chat template for a 2-turn conversation. For a given generation, only reasoning content from the current turn is materialized into the prompt.}
    \label{fig:chat_template}
\end{figure}

\subsubsection{Data}

\textbf{Competition Math.} For math, we use a similar strategy to Nemotron Nano 2 \citep{nemotronnanov2}. However, we refresh the responses with GPT-OSS 120B \citep{openai2025gptoss120bgptoss20bmodel}. In addition, we create tool-integrated reasoning traces using Python tools and GPT-OSS 120B as the teacher model.

\textbf{Competition Code.} For code we use the same data from Nemotron Nano 2, which is made up of the prompts from \citet{ahmad2025opencodereasoning} complemented with responses from DeepSeek-R1-0528 \citep{deepseekai2025deepseekr1incentivizingreasoningcapability}.

\textbf{Conversational Tool Use.} We generate synthetic multi-turn trajectories to demonstrate conversational tool use. The generation of these trajectories involves a user that is given a task to accomplish, an agent that is instructed to help the user accomplish their task, and a tool execution environment, each of which is simulated by a language model. To limit the trajectories in the SFT training data to ones in which the actions of all of these entities are consistent with their goals, we employ a language model as a judge to evaluate the trajectories, and filter out trajectories for which the judge considers an action of an entity to be inconsistent with its goals. We use Qwen3-235B-A22B-Thinking-2507 \citep{yang2025qwen3technicalreport}, Qwen3-32B \citep{yang2025qwen3technicalreport}, GPT-OSS-120b \citep{openai2025gptoss120bgptoss20bmodel}, and Qwen3-235B-A22B-Instruct-2507 \citep{yang2025qwen3technicalreport} to generate data in this synthetic tool use trajectory generation pipeline.

\textbf{Long Context.} We generate synthetic data with a mean token length of 128k tokens and a maximum of 256k tokens to improve long-context performance, validated against a subset of RULER tasks.

\textbf{Formal Proofs.} For Lean theorem proving, we curated SFT data by first autoformalizating ~580k natural language theorems from online mathematics communities (AoPS, Math StackExchange, MathOverflow) into ~550k Lean 4 statements using an iterative refinement pipeline based on GPT-OSS-120B with backtranslation-based semantic verification. We then ran large-scale proof generation using Goedel-Prover-V2-32B with up to 4 independent attempts and 8 self-correction rounds per statement, yielding ~920k proof traces with compiler-verified solutions. After filtering, the final dataset contains ~300k examples pairing formal theorem statements with successful reasoning traces and proofs.

\textbf{Multilingual.} We generate multilingual data in a similar manner to Nemotron Nano 2 \citep{nemotronnanov2}. We used Qwen$2.5$-Instruct to translate our existing English post-training data into $5$ target languages, French, Spanish, Italian, German and Japanese. Our pipeline translates inputs line-by-line and skips non-translatable content like code, XML tags, URLs, etc. 
Following the translation of the English source text, we utilized a language identification tool \url{https://pypi.org/project/langdetect/} to filter out samples that did not predominantly consist of target language tokens. Additionally, we excluded samples containing specific failure modes where the Qwen model explicitly stated its inability to translate the source text.

Our multilingual corpus was further comprised of $1.62$ million text translation samples, aggregated from a combination of news-commentary datasets and proprietary sources. These samples covered bidirectional translation tasks between English and the five target languages.

\textbf{Terminal Use.} To teach \ourmodel to complete autonomous terminal-based tasks, we generate a diverse set of verifiable tasks based on Terminal Bench \citep{tbench_2025}. In particular, we adapt data from our competitive coding, competitive math, and long context datasets to terminal bench problems. We also constructed synthetic tasks requiring data analysis and file creation and operations. Additionally, we incorporated data from SWE-Smith \citep{swesmith}, which provides real-world software engineering tasks. We use 
Qwen3-Coder-480B-A35B-Instruct \citep{qwen2025qwen25technicalreport} and Kimi-K2-Instruct-0905 \citep{kimiteam2025kimik2openagentic} to generate action trajectories for each task using the Terminus-1 and Terminus-2 agents \citep{tbench_2025}.

\textbf{General Chat.} We create SFT data by generating responses to the LMSYS \citep{zheng2023judging} and WildChat datasets \citep{li2024wildchat} using GPT-OSS-120B, Qwen3-235B-A22B-Thinking-2507, and Qwen3-235B-A22B-Instruct-2507. The data is extended to multi-turn by having the same language model simulate the user and further continue the conversation.

\textbf{Instruction Following.} We create targeted instruction following data with the methodology used in Tülu 3 \citep{lambert2025tulu3pushingfrontiers}. We simulate users in a conversation using language models seeded with a user persona from Nemotron-Personas-USA \citep{nvidia/Nemotron-Personas-USA} and instructions from IFeval \citep{zhou2023instruction} and IFBench \citep{pyatkin2025generalizing} train splits. The user language model is prompted to generate precise instruction following queries for one or more turns. We then use GPT-OSS-120B, Qwen3-235B-A22B-Thinking-2507, and Qwen3-235B-A22B-Instruct-2507 to generate responses to the user queries. The generated data is first filtered to only keep samples where all turns pass the respective instruction verifier implementations in IFEval and IFBench. Further filtering is done with a language model judge to remove samples where the responses only trivially or superficially follow instructions.

\textbf{Safety.} 
We compile a diverse set of unsafe prompts sourced from the Nemotron Content Safety v2 \citep{ghosh-etal-2025-aegis2} and the Gretel Safety Alignment v1 \citep{gretelai_gretel-safety-alignment-en-v1}  datasets to target content safety risks, and Harmful Tasks \citep{hasan2024pruning} and Red-Team-2K \citep{luo2024jailbreakv_robustness} datasets to target common jailbreak techniques. This collection is further balanced with safe prompts derived from Nemotron Content Safety v2.

For supervised fine-tuning (SFT), we apply safe prompt wrappers to unsafe prompts enabling the models to learn appropriate refusal behaviors while preserving user engagement. Various refusal strategies are implemented to align with good user experience. For instance, self-harm related prompts are paired with prompt templates encouraging the use of appropriate suicide prevention helplines. A content-safety classifier is employed to filter the responses, ensuring alignment with safety objectives.


\textbf{Software Engineering.} To train \ourmodel for autonomous software engineering capabilities including code exploration, issue reproduction and bug fixing, we curate a dataset of coding tasks derived from real-world GitHub issues. We use the issue description and containerized execution environments from SWE-Gym \citep{pan2025trainingsoftwareengineeringagents} and R2E-Gym \citep{jain2025r2egymproceduralenvironmentshybrid} datasets. We distill trajectories from three open-source agent harnesses - OpenHands \citep{wang2025openhandsopenplatformai}, SWE-Agent \citep{yang2024sweagent}, and Mini-SWE-Agent \citep{yang2024sweagent} using Qwen3-Coder-480B-A35B-Instruct  \citep{qwen2025qwen25technicalreport} as the teacher model. 

\textbf{Science.} The science dataset spans physics, chemistry, and biology, and is produced through a unified pipeline that integrates synthetic, real, and document-based seed sources. We began by curating a set of challenging seed questions derived from Nemotron Nano v2 \citep{nemotronnanov2} as well as from scientific articles contained in the pre-training corpus. In parallel, we incorporated additional scientific articles from the same corpus as a complementary reservoir of seed material. Each article was annotated with three attributes: (1) education domain based on bert-based finetuned classifier \citep{li2024datacomp}, (2) content level (ranging from elementary to graduate), and (3) fine-grained topical categories (e.g., biology, chemistry, mathematics, law). Focusing on the graduate-level subset, we indexed these documents in a vector database and used a diverse set of science-oriented query prompts to retrieve thousands of highly relevant passages. These retrieved segments served as the foundation for generating multiple-choice question (MCQ) data, which were subsequently converted into an open-ended question-answering (OpenQA) format.

All seed sources—synthetic, real, and doc-retrieved—were subsequently processed through NeMo Data Designer \citep{nemo-data-designer}. The Data Designer was used to paraphrase prompts, produce multiple format and instruction variants, and enhance robustness across prompt styles. Reasoning traces for the SFT stage were generated using tool-integrated Python reasoning traces from GPT-OSS 120B \citep{openai2025gptoss120bgptoss20bmodel}. Crucially, all generated variants underwent rigorous LLM-judge filtering, ensuring strict format compliance, intent preservation, and high-quality reasoning consistency. During the RL stage, we further introduced targeted prompt and format augmentations to reduce prompt sensitivity and improve generalization.

A subset of STEM datasets developed in this work are released in both the multiple-choice question (MCQ\footnote{\url{https://huggingface.co/datasets/nvidia/Nemotron-RL-knowledge-mcqa}}
) and open question-answering (OpenQA\footnote{\url{https://huggingface.co/datasets/nvidia/Nemotron-RL-knowledge-openqa}}
) formats to support Nano-V3 training and broader downstream research. These datasets are fully integrated into the RLVR pipeline, with both MCQ\footnote{\url{https://github.com/NVIDIA-NeMo/Gym/tree/main/resources_servers/mcqa}}
 and OpenQA\footnote{\url{https://github.com/NVIDIA-NeMo/Gym/tree/main/resources_servers/equivalence_llm_judge}}
 environments provided through NeMo Gym \citep{nemo-gym}. This unified pipeline ensures consistent quality standards and supports robust reinforcement-learning-based evaluation and training across all STEM domains.

\textbf{GenSelect.} We improve our model’s capability as a generative reward model by training it to identify the best solution among multiple candidates, following the approach in \cite{toshniwal2025genselectgenerativeapproachbestofn}. We adapted the problems in our math and coding SFT data by generating synthetic solutions and then selection reasoning traces including their final verdicts using GPT-OSS 120B \citep{openai2025gptoss120bgptoss20bmodel} and DeepSeek-R1-0528 \citep{deepseekai2025deepseekr1incentivizingreasoningcapability}.  

\textbf{CUDA.} We collect and synthesize 21k (PyTorch, Cuda C) pairs with seeds from HuggingFace Transformers \citep{wolf-etal-2020-transformers} and KernelBook \citep{kernelbook2025}. We first parse the PyTorch code from Transformers \citep{wolf-etal-2020-transformers} and KernelBook \citep{kernelbook2025}, and then use DeepSeek-R1-0528 \citep{deepseekai2025deepseekr1incentivizingreasoningcapability} to generate corresponding Cuda C code. We only include {PyTorch, Cuda C} pairs with Cuda C code that is successfully compiled and numerically verified against PyTorch reference code. 

\subsubsection{Data Filtering}
For all domains, we apply a unified data filtering pipeline to ensure that only high-quality, license-compliant, and verifiable samples are used for training. We first discard malformed examples using structural checks (e.g., missing tool definitions when tool calls are present). We then aggressively filter reasoning traces exhibiting pathological repetition, such as repeated n-grams within a sliding window or across the entire trajectory, which we found to be a strong indicator of malformed or low-quality reasoning. Finally, based on internal audits of synthetically generated datasets, we observed that some teacher models occasionally produce reasoning traces and final responses that implicitly align with specific political entities or promote nationalistic narratives. To mitigate this, we apply targeted keyword- and regex-based filters (e.g., patterns such as “our {nation/party} […]”, “our values”) and remove all trajectories matching such behavior.

\subsubsection{Data Mixture}
\begin{figure}[t]
    \centering
    \includegraphics[width=10cm]{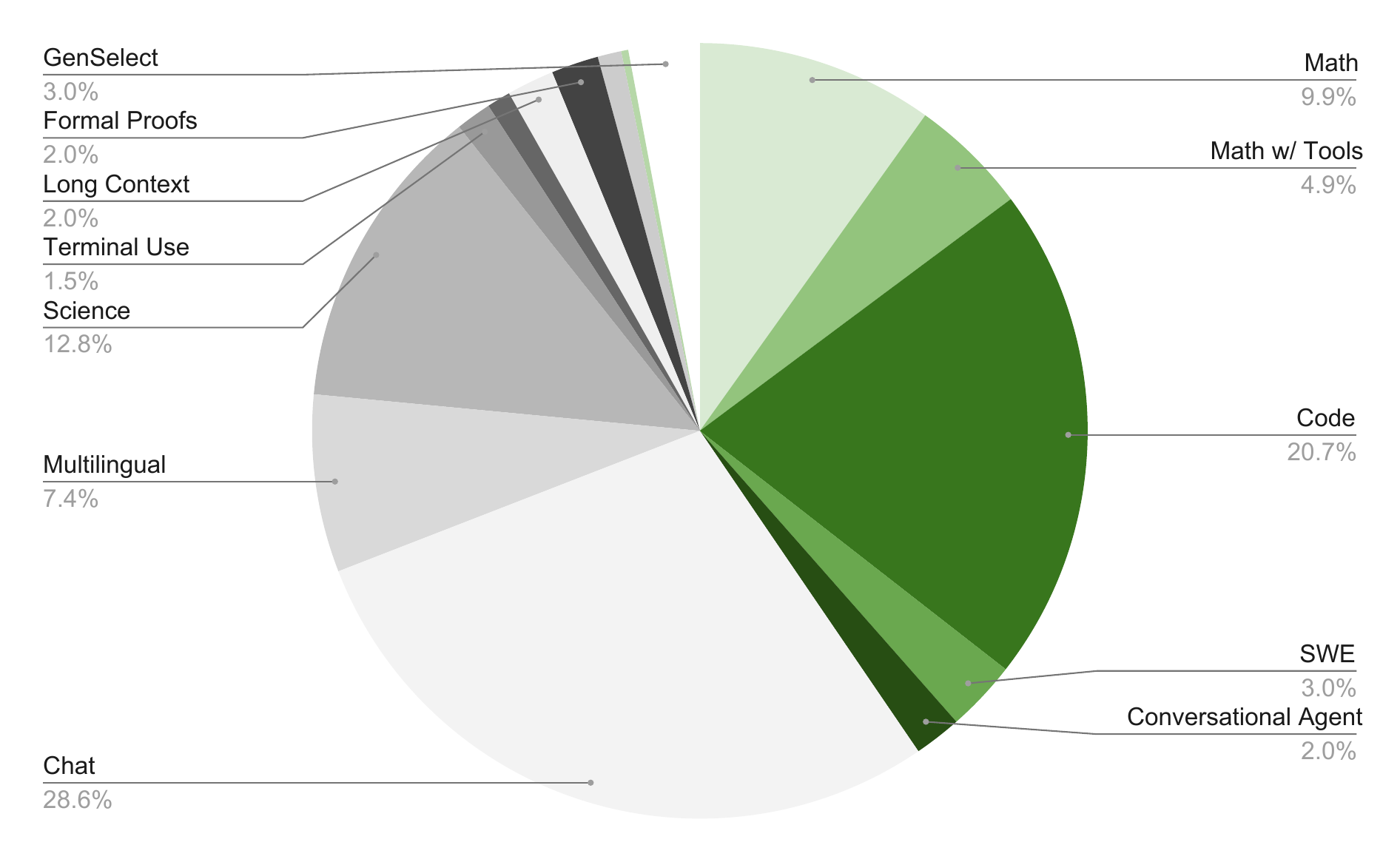}
    \caption{SFT data blend for \ourmodel. }
    \label{fig:sft_blend}
\end{figure}

Our exact data blend can be found in Figure \ref{fig:sft_blend} (all datasets not listed make up less than 1\% of the blend). We train over 18M total samples. For each dataset we decide how much data to include based on the approximate amount of data required to achieve optimal performance in single task settings. As the size of different datasets varies significantly, we employ a dynamic sampling approach where smaller datasets may be trained over for many epochs and larger datasets are trained for only a few epochs.

\subsubsection{Reasoning Control}
\ourmodel allows for two different forms of reasoning control: reasoning on/off control and token budget control. Similar to \citet{nemotronnanov2}, to enable reasoning on/off control we strip the reasoning traces from a random 10\% of samples, and to enable budget control, we randomly truncate 3\% of reasoning traces to different reasoning budgets, before continuing with the original post-reasoning response.

\subsubsection{Hyperparameters}
We train for 13000 steps using a batch size of 64 and employ sequence packing to a sequence length of 256K. We use a learning rate of $5\cdot 10^{-5}$ and use 800 steps of learning rate warmup. We use a sequence-level MoE load balancing regularizer and set the loss coefficient to $10^{-4}$.


\subsection{Multi environment Reinforcement Learning from Verifiable Rewards}
\label{subsec:RLVR}
We employ a unified RLVR stage, training on all environments simultaneously. We find that this results in stable gains across all benchmarks throughout training, while single environment training often results in un-recoverable degradation of other benchmarks. We do two stages of such RLVR: one immediately after SFT and one after RLHF.
\subsubsection{Environments}


\textbf{Competition Math.} We train on the DAPO \citep{yu2025dapo} and SkyWorks math \citep{he2025skywork} datasets. These datasets have 17K and 104K tasks respectively.

\textbf{Competition Coding.} We use competitive coding problems from \citet{ahmad2025opencodereasoning}. We limit the number of unit tests to 50 in order to reduce verification time. This filtering leaves us with 22K tasks.

\textbf{Question Answering.} We train on a variety of difficult multiple choice datasets focusing on STEM domains. Here the questions and answers are generated based on information from reference documents. This dataset has with 135K tasks.

\textbf{Structured Outputs.} We train \ourmodel to have strong JSON schema adherence capabilities. We utilized NeMo Data Designer \citep{nemo-data-designer} to create the seed dataset for RL. We start by constructing (JSON schema, document) pairs conditioned on diverse topics using Qwen3-235B-A22B-Instruct-2507 \citep{yang2025qwen3technicalreport}. We then utilized these pairs to create RL prompts by taking the model to summarize the document according to the schema.  To ensure high syntactic validity, the pipeline enforced strict complexity controls and applied rejection sampling, while simultaneously varying instruction difficulty and phrasing to maximize input diversity. This pipeline produces 9K tasks.

In the RL stage, a positive reward is given when the output matches the exact schema constraints, and no reward is given otherwise. For simplicity, we do not add a reward for the semantic content of the output.

\textbf{Instruction Following.} We use two instruction following environments during the training. The first environment is similar to the IFEval style environment used in \citet{bercovich2025llamanemotronefficientreasoningmodels}, but with refreshed constraints from the IFBench training set \citep{pyatkin2025generalizing}. We create 46K tasks for this environment.

The second environment uses LLM as a judge to verify whether or not the agent has followed complex instructions in multi-turn settings, where the instructions may be quite subtle. This environment is inspired by the Multi-Challenge benchmark \citep{deshpande2025multichallenge}. We create 3K total tasks for it.

\textbf{Long Context.}
We generate challenging long-context QA pairs using Qwen3-235B-A22B-Thinking-2507 \citep{yang2025qwen3technicalreport}, drawing from a subset of our pre-training mixture designed for multi-document synthesis. Each question is required to reference at least five documents, with the total input limited to 32k tokens. We employ Qwen3-235B-A22B-Instruct-2507 \citep{yang2025qwen3technicalreport} as the LLM judge to evaluate the model's rollouts. This dataset contains 12K tasks.

\textbf{Agentic Tool Use.} We use two environments to improve tool use capabilities. The first is Workplace Assistant, a multi-step verifiable tool-calling setup adapted from Styles \citep{Styles2024WorkBench} that was also used in Nemotron 2 Nano \citep{nemotronnanov2}. This is a tool use - multi step agentic environment that tests the agent's ability to execute tasks in a workplace setting. Workplace Assistant contains a sandbox environment with five databases, 26 tools, and 690 tasks. These tasks represent common business activities, such as sending emails, scheduling meetings, etc. The correctness is verified through executing the tool calls issued by the agent and comparing it to the ground truth database state. 

The second environment is a Multi-turn conversational agent environment. It tests an agent's tool-calling and proactive asking capability. Comprising approximately 1K tasks, this environment simulates complex banking scenarios like assisting customers with unblocking a credit card or solving account disputes. The correctness of the agent's actions is automatically verified by executing the tool calls it issues and comparing the resulting database state against the predefined ground truth.

\subsubsection{Data Mixture and Curriculum}
We begin by profiling all reinforcement learning (RL) tasks using our supervised fine-tuning (SFT) checkpoint. To focus training on challenging cases, we filter out samples where the SFT checkpoint already achieves a 100\% pass rate. We then adopt the curriculum training method introduced in~\citet{bercovich2025llamanemotronefficientreasoningmodels}, which dynamically adjusts task difficulty throughout training.

In each batch, we maintain a fixed ratio of samples across different domains. For each domain, we model the target pass-rate distribution as a Gaussian function, shifting from high pass-rate (easier) samples early in training to low pass-rate (harder) samples later. The target mean of Gaussian distribution decreases linearly throughout training steps. Within each batch, samples from different domains are shuffled. This Gaussian sampling strategy prevents overfitting to either overly easy or overly difficult examples, ensuring a balanced learning progression. 

\begin{figure}[ht]
    \centering
    \includegraphics[width=0.6\textwidth]{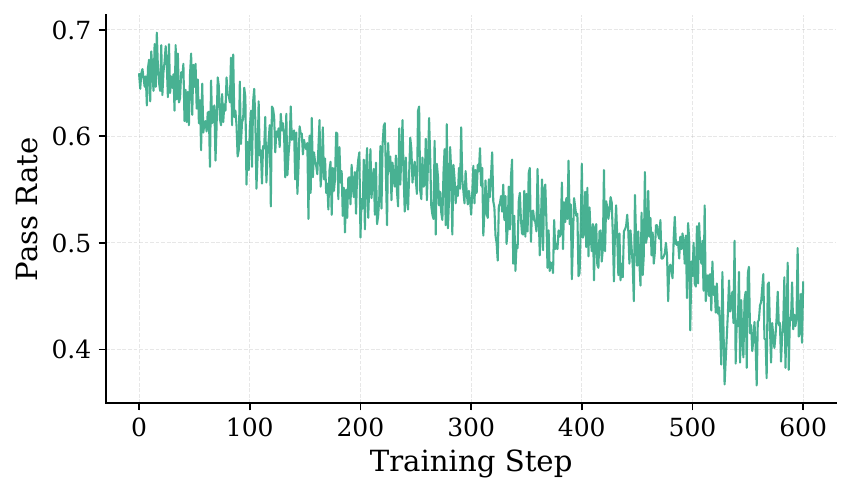}
    \caption{Batch-wise pass rates across the RL curriculum.}
    \label{fig:pass_rate}
\end{figure}

This approach enables a controlled and gradual increase in task difficulty while preserving domain diversity and ensuring efficient batch composition. Figure~\ref{fig:pass_rate} illustrates how sample difficulty evolves over the course of RL training. Once training progress plateaus, we re-profile the tasks using the best RL checkpoint and construct a new curriculum to further refine performance.

\begin{figure}[ht]
    \centering
    \includegraphics[width=\textwidth]{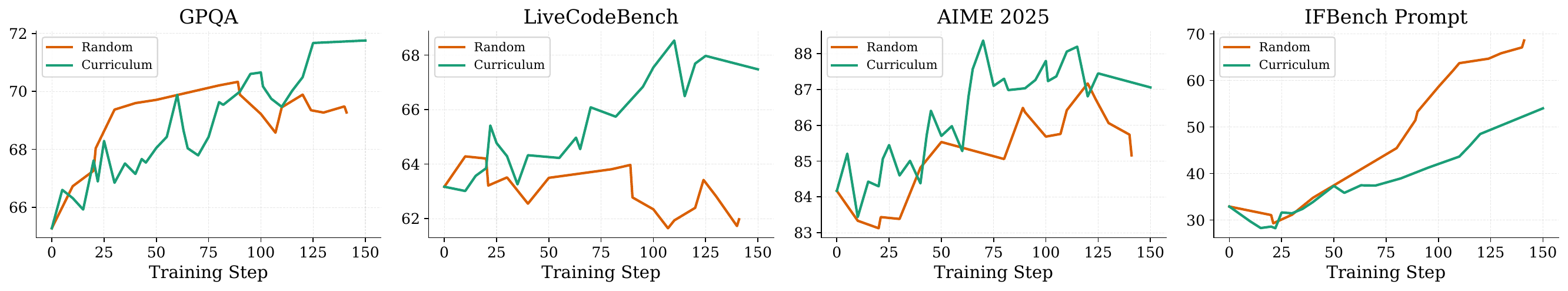}
    \caption{Comparison between curriculum sampling and random sampling.}
    \label{fig:curriculum_ablation}
\end{figure}

We compare curriculum sampling against random sampling using an intermediate SFT checkpoint, maintaining identical domain ratios in both cases. As shown in Figure~\ref{fig:curriculum_ablation}, curriculum sampling ensures stable learning across multiple domains throughout training. In contrast, random sampling biases the model toward easier tasks, preventing it from effectively learning more challenging ones.

\subsubsection{Surpassing SFT with RLVR}

\begin{figure}[ht]
    \centering
    \includegraphics[width=\textwidth]{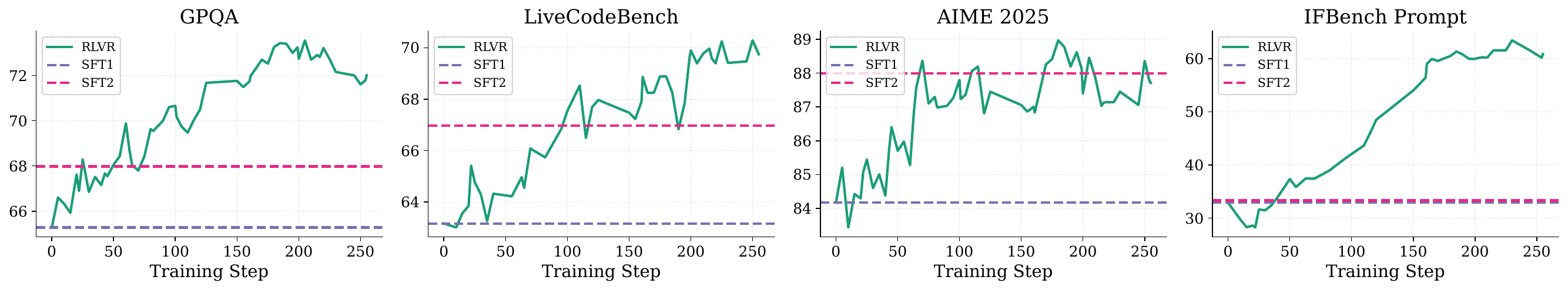}
    \caption{RLVR surpasses or matches heavily fine-tuned SFT model across all evaluated domains.}
    \label{fig:surpass_sft_with_rlvr}
\end{figure}

Recent works have demonstrated that supervised fine-tuning (SFT) alone on small models can achieve strong performance~\citep{ahmad2025opencodereasoning,deepseekai2025deepseekr1incentivizingreasoningcapability}. In this study, we investigate whether RLVR can outperform a heavily fine-tuned SFT baseline. As illustrated in Figure~\ref{fig:surpass_sft_with_rlvr}, we compare the accuracy of model during RLVR training with two SFT checkpoints: 
\begin{itemize} 
\item SFT1: Our initial RLVR starting point, fine-tuned for approximately 3 epochs. 
\item SFT2: A heavily fine-tuned checkpoint, trained to full convergence (approximately 5 epochs). \end{itemize} 
Our results show that even with relatively short training, RLVR consistently exceeds  or matches the accuracy of the heavily fine-tuned SFT model across all evaluated domains.

\subsubsection{Infrastructure}
\label{subsubsection:RLinfra}

RL at the frontier of model post-training is currently defined by scaling up to an increasing diversity of tasks or environments designed for the model to learn increasingly general capabilities.
Scaling RL to many environments requires a high-performance, extensible, and standardized interface for coordinating between rollouts and training.
To address the scaling performance and extensibility challenges using one standard framework, we adopt NeMo Gym \citep{nemo-gym} and NeMo RL \citep{nemo-rl} for enabling large-scale RL on many different environments/verifiers.

NeMo Gym is based on the abstraction of \emph{servers}.
There are three core varieties of servers in Gym: (1) \emph{agents}, (2) \emph{models}, and (3) \emph{resources}.
An \emph{agent} server implements the rollout kernel of a RL environment.
A \emph{model} server wraps an inference engine such as vLLM \citep{kwon2023efficientmemorymanagementlarge} to provide a prompt-response API, and also carefully preserves token and inference log-prob data and metadata required for RL.
A \emph{resource} server provides a verification API for computing rewards from a given rollout.

Our Nemotron Nano 3 RLVR experiments were all based on an integrated infrastructure of NeMo RL and NeMo Gym:
NeMo RL acts as the RL training loop controller, using Megatron-Core \citep{shoeybi2020megatronlmtrainingmultibillionparameter} for model training at scale, and routing all rollouts through NeMo Gym and vLLM.

\subsubsection{Algorithm}
\label{sec:rl-algo}
We train \ourmodel using synchronous GRPO with masked importance sampling to mitigate training-inference misalignment \citep{Shao2024DeepSeekMath, team2025every, yao2025offpolicy}. We use 128 prompts per step and use 16 generations per prompt. We train with a batch size of 2048, making our updates on-policy. To further stabilize training we also freeze the MoE router weights. We employ the aux-loss-free load balancing approach and keep updating expert bias~\citep{wang2024auxiliary}. 

Our entire training run is done with a maximum generation length of 49K. We use overlong filtering \citep{yu2025dapo}, which we find boosts performance on reasoning intensive benchmarks.


\begin{figure}[ht]
    \centering
    \includegraphics[width=\textwidth]{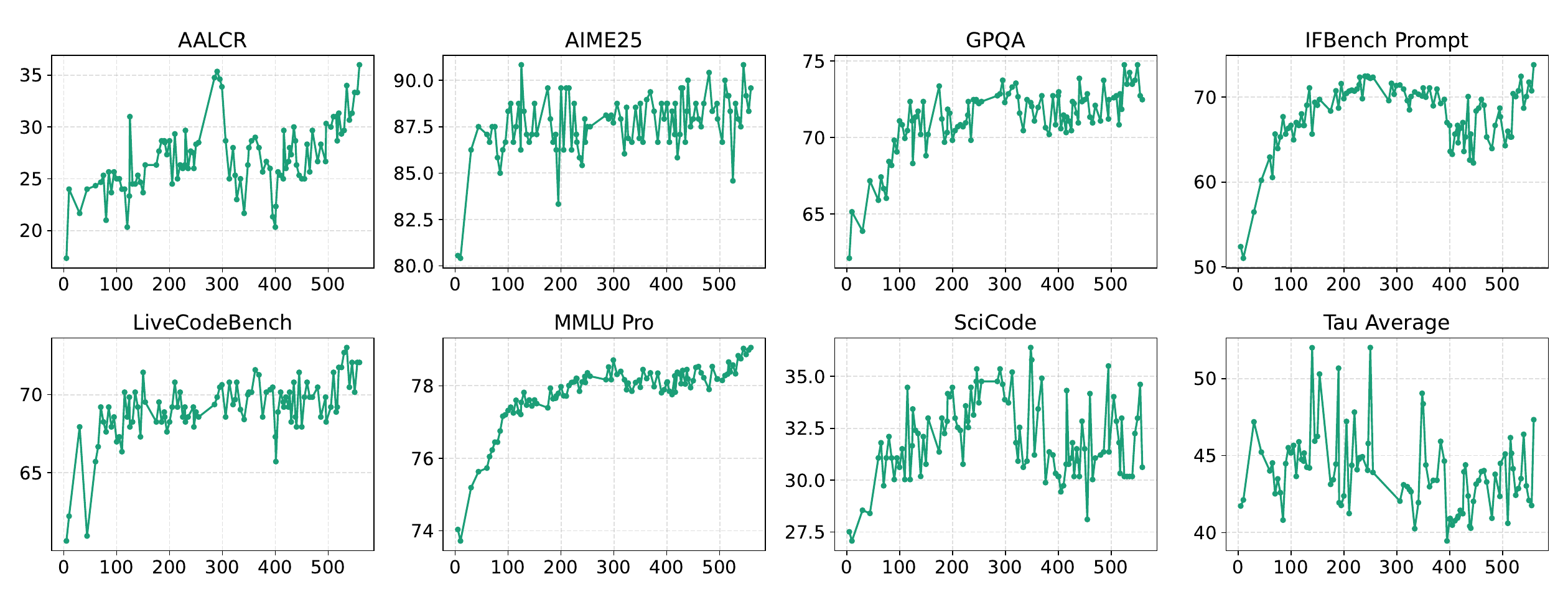}
    \caption{Benchmark performance throughout RL training.}
    \label{fig:rl_progress}
\end{figure}
\subsection{Reinforcement Learning from Human Feedback}
\label{subsec:RLHF}

\subsubsection{Scaling Reinforcement Learning for Generative Reward‑Model Training}
Many recent works \citep{deepseek-genrm,helpsteer3-pref,rmr1} have demonstrated that generative reward models (GenRMs) generalize better than traditional Bradley‑Terry models, reducing the risk of reward hacking during RLHF. In order to train an accurate and robust GenRM, we leverage reinforcement learning at scale. Building on the methodology of \citet{helpsteer3-pref}, we train Qwen3-235B-A22B-Thinking-2507~\citep{yang2025qwen3technicalreport} to become a GenRM with GRPO algorithm.  Given the conversation history, a new user request, and two candidate assistant responses, the GenRM first reasons through the strength and weakness of both responses, then produce an individual helpfulness score for each response as well as a ranking score. For GenRM training, we use 128 prompts per batch, 8 generations per prompt, and do one gradient step on the full batch. We define the reward as

\begin{align}
    \mathbf{R} = -C_1I_{\text{format}} -
    \left| P_{h1} - G_{h1} \right| - \left| P_{h2} - G_{h2} \right| - C_2 \left| P_r - G_r \right|,
\end{align}

where $P_r$, $G_r$ denote the predicted and ground-truth preference rankings; $P_{h1}$, $G_{h1}$, $P_{h2}$, $G_{h2}$ denote the predicted and ground-truth helpfulness scores for responses 1 and 2, respectively;  $I_{format}$ indicates whether the prediction violates the format requirement; $C_1$ and $C_2$ are hyper-parameters controlling the weights. We set $C_1=10$ and $C_2=1$. 

\begin{figure}[ht]
    \centering
    \includegraphics[width=\textwidth]{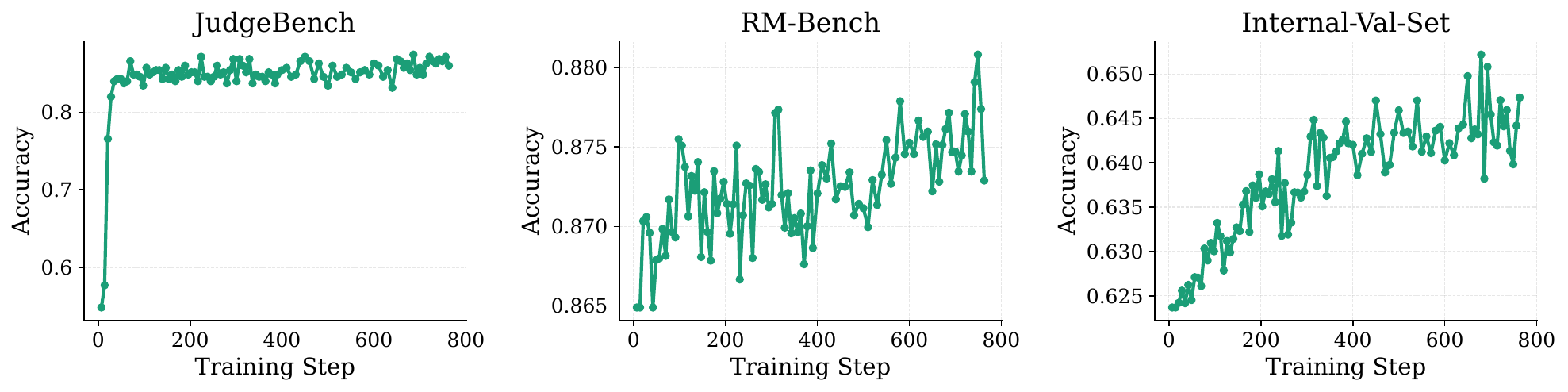}
    \caption{GenRM performance improves across benchmarks as we scale up RL training.}
    \label{fig:genrm-training}
\end{figure}

We leverage data from HelpSteer3~\citep{helpsteer3-pref}, a commercially-friendly subset of lmarena-ai/arena-human-preference-140k~\citep{chiang2024chatbot}, and a synthetic safety blend (see details in Appendix~\ref{sec:safety_pref_data}) for model training. In our dataset, individual helpfulness scores range from 1 to 5, where higher means more helpful, while ranking score ranges from 1 to 6, in which 1 denotes that response 1 is far superior to response 2 and 6 denotes that response 2 is far superior to response 1~\citep{helpsteer3-pref}. We augment each sample by switching positions of two responses to prevent positional bias. Figure~\ref{fig:genrm-training} demonstrates that the performance of GenRM on RM-Bench~\citep{rmbench}, JudgeBench~\citep{judgebench}, and our internal validation set steadily improves as training progresses.

\subsubsection{RLHF with Group Relative Length Control}
With a trained GenRM, we conduct RLHF on the same set of prompts. Same as RLVR, we use a batch of 128 prompts and 16 responses per prompt. Naively comparing all pairs of $N$ responses would require $\binom{N}{2}$ GenRM calls per prompt, which scales quadratically and becomes prohibitively expensive for large $N$. With $N=16$ responses, this would require 120 comparisons per prompt. Instead, we adopt a circular comparison strategy where each response is compared only with its successor: $(r_1, r_2), (r_2, r_3), \ldots, (r_{N-1}, r_N), (r_N, r_1)$, yielding exactly $N$ comparisons. This reduces computational cost from $O(N^2)$ to $O(N)$ while still connecting all responses in a comparison graph. Each response is also judged twice in different positions so as to alleviate positional bias.

For each pairwise comparison $(r_i, r_j)$, the GenRM produces individual helpfulness scores $s_i, s_j \in [1, 5]$ and a ranking score $s_r \in [1, 6]$. In the case where $s_i=s_j$, we further employ a simple tiebreaker mechanism:

\begin{align}
   s_i=s_i+(3.5-s_r), \\
   s_j = s_j + (s_r-3.5).
\end{align}

The base reward $R_i^{(\text{base})}$ for response $r_i$ is then computed by averaging its scores from two matches.

When training with base reward, we find that the length of response can rapidly increase as RLHF training proceeds. This is different from reward hacking, as the increase of length mostly comes from reasoning trace while only final answer is judged by GenRM. It is similar to observations in \citet{deepseekai2025deepseekr1incentivizingreasoningcapability} where model spends more inference time compute to achieve better rewards. However, unlike reasoning heavy tasks like math and coding, prompts in RLHF datasets usually don't require extensive reasoning. In order to reduce redundant thinking, we propose a Group Relative Length Control mechanism during RLHF. Specifically, for each prompt, we generate a group of $N$ candidate responses $\{r_1, r_2, \ldots, r_N\}$. Each response $r_i$ is decomposed into a reasoning component $r_i^{(\text{think})}$ and an answer component $r_i^{(\text{answer})}$, with corresponding lengths $\ell_i^{(\text{think})}$ and $\ell_i^{(\text{answer})}$.

\paragraph{Length-Normalized Reward Adjustment.} We compute a zero-mean, group-relative length bonus that encourages shorter responses within a group. For the reasoning component, we first normalize lengths within the group
\begin{align}
    w_i^{(\text{think})} = 1 - \frac{\ell_i^{(\text{think})} - \ell_{\min}^{(\text{think})}}{\ell_{\max}^{(\text{think})} - \ell_{\min}^{(\text{think})}},
\end{align}

where $\ell_{\min}^{(\text{think})} = \min_j \ell_j^{(\text{think})}$ and $\ell_{\max}^{(\text{think})} = \max_j \ell_j^{(\text{think})}$. To ensure the adjustment is zero-sum across the group (preserving the overall reward scale), we center the weights
\begin{align}
    \tilde{w}_i^{(\text{think})} = w_i^{(\text{think})} - \frac{1}{N}\sum_{j=1}^{N} w_j^{(\text{think})}.
\end{align}

The same procedure is applied to answer lengths to obtain $\tilde{w}_i^{(\text{answer})}$. The final reward for response $r_i$ is then
\begin{align}
    R_i = R_i^{(\text{base})} + \lambda^{(\text{think})} \tilde{w}_i^{(\text{think})} + \lambda^{(\text{answer})} \tilde{w}_i^{(\text{answer})},
\end{align}

where $R_i^{(\text{base})}$ is the base reward from pairwise comparisons and $\lambda^{(\text{think})}, \lambda^{(\text{answer})}$ are  coefficients controlling the strength of the length penalty. We set $\lambda^{(\text{think})}=0.5$, $\lambda^{(\text{answer})}=0.5$.
\paragraph{Quality-Gated Conciseness Bonus.} To further encourage concise responses without sacrificing quality, we introduce optional bonuses for the shortest responses that achieve top-tier quality scores. Let $\tau_p$ denote the $p$-th percentile threshold of scores within the group. For the response $r_k$ with minimum reasoning length:
$$R_k \leftarrow R_k + \beta^{(\text{think})} \cdot \mathbb{1}\left[R_k^{(\text{base})} \geq \tau_p\right]$$
Similarly, for the response $r_m$ with minimum answer length:
$$R_m \leftarrow R_m + \beta^{(\text{answer})} \cdot \mathbb{1}\left[R_m^{(\text{base})} \geq \tau_p\right]$$
where $\beta^{(\text{think})}$ and $\beta^{(\text{answer})}$ are the reasoning and answer conciseness bonuses respectively, and $\mathbb{1}[\cdot]$ is the indicator function. We set $\beta^{(\text{think})}=0.5$, $\beta^{(\text{answer})}=0.5$, and $\tau_p=80$.

This mechanism ensures that (1) length penalties are relative within each prompt group rather than absolute, avoiding bias against inherently complex problems; and (2) conciseness bonuses are only awarded to high-quality responses, preventing the model from learning to produce short but low-quality answers. We observe that the verbosity level reduces 30\% during the training without sacrificing accuracy.

\subsection{Post-trained Model Evaluations}
\label{subsec:final_model_evals}

\subsubsection{Evaluation Benchmarks}

\begin{table*}[!htp]
\centering
\small
\setlength{\tabcolsep}{7pt}
\renewcommand{\arraystretch}{1.15}

\begin{tabular}{l|c c c }
\toprule
\textbf{Benchmark} & \textbf{N-3-Nano} & \textbf{Qwen3} & \textbf{GPT-OSS} \\
\midrule

\rowcolor{black!5}
\multicolumn{4}{l}{\textbf{General Knowledge}} \\
MMLU-Pro &  78.30 &  \textbf{80.90} & 75.00 \\

\midrule
\rowcolor{black!5}
\multicolumn{4}{l}{\textbf{Reasoning}} \\
AIME25 (no tools) & 89.06 & 85.00 & \textbf{91.70} \\
AIME25 (with tools) & \textbf{99.17} & - & 98.7 \\
GPQA (no tools) &  73.04 & \textbf{73.40} & 71.50 \\
GPQA (with tools) & \textbf{75.00} & - & 74.20 \\
LiveCodeBench (v6 2024-08$\leftrightarrow$2025-05) & \textbf{68.25} & 66.00 & 61.00\\
SciCode (subtask) & 33.28 & 33.00 & \textbf{34.00}  \\
HLE (no tools) & 10.57 & 9.80 & \textbf{10.90} \\
HLE (with tools) & 15.48 & - & \textbf{17.30} \\
MiniF2F pass@1 & \textbf{50.03} & 5.72* & 12.05* \\
MiniF2F pass@32 & \textbf{79.92} & 16.80* & 43.03* \\
\midrule
\rowcolor{black!5}
\multicolumn{4}{l}{\textbf{Agentic}} \\
Terminal Bench (hard subset) & 8.51 & 5.00 & \textbf{10.00} \\
SWE-Bench (OpenHands) & \textbf{38.76} & 22.00* & 34.00* \\
\textbf{TauBench V2} &  &  &  \\
\quad Airline & 48.00 & \textbf{58.00} & 38.00 \\
\quad Retail & 56.91 & \textbf{58.80} & 54.80 \\
\quad Telecom & 42.21 & 26.30 & \textbf{49.70} \\
\quad Average & \textbf{49.04} & 47.70 & 47.50 \\
BFCL v4 & \textbf{53.76} & 46.40* & - \\
\midrule
\rowcolor{black!5}
\multicolumn{4}{l}{\textbf{Chat \& Instruction Following}} \\
IFBench (prompt) & \textbf{71.51} & 51.00 & 65.00 \\
Scale AI Multi Challenge & 38.45  & \textbf{44.75} & 33.75 \\
Arena-Hard-V2 (Hard Prompt) & \textbf{72.10} & 49.60* & 71.20* \\
Arena-Hard-V2 (Creative Writing) & 63.20 &  \textbf{66.00*} & 25.90* \\
Arena-Hard-V2 (Average) & \textbf{67.65} & 57.80 & 48.55 \\

\midrule
\rowcolor{black!5}
\multicolumn{4}{l}{\textbf{Long Context}} \\
AA-LCR & 35.85 & \textbf{59.00} & 34.00 \\
RULER-100 @ 256k & \textbf{92.92} & 89.40 & - \\
RULER-100 @ 512K & \textbf{91.25} & 84.00 & - \\
RULER-100 @ 1M & \textbf{86.34} & 77.50 & - \\

\midrule
\rowcolor{black!5}
\multicolumn{4}{l}{\textbf{Multilingual}} \\
MMLU-ProX (avg over langs) & 59.50 & \textbf{77.60*} & 69.10* \\
WMT24++ (en$\rightarrow$xx) & \textbf{86.20} & 85.60 & 83.20 \\

\bottomrule
\end{tabular}

\caption{
\ourmodel compared to \text{Qwen3-30B-A3B-Thinking-2507}, and \text{GPT-OSS 20B}. 
}
\label{tab:nano_v3_comparison}
\end{table*}

We evaluate \ourmodel across a broad suite of established benchmarks spanning mathematical and
scientific reasoning, coding, agentic tool use, instruction following, long-context understanding, and
multilingual capability. Table~\ref{tab:nano_v3_comparison} summarizes the final results.

All evaluation results were collected via Nemo Evaluator SDK\footnote{\url{https://github.com/NVIDIA-NeMo/Evaluator}} and for most benchmarks, the Nemo Skills Harness\footnote{\url{https://github.com/NVIDIA-NeMo/Skills}}. For reproducibility purposes, the open source container on Nemo Skills packaged via NVIDIA's Nemo Evaluator SDK used for evaluations can be found here\footnote{\url{https://catalog.ngc.nvidia.com/orgs/nvidia/teams/eval-factory/containers/nemo_skills}}. In addition to Nemo Skills, the evaluations also used dedicated packaged containers for Tau-2 Bench, ArenaHard v2, AA\_LCR. More details on the evaluation settings can be found in the Nemo Evaluator SDK configs folder\footnote{\url{https://github.com/NVIDIA-NeMo/Evaluator}}. The following benchmarks are not onboarded yet in our open source tools and for these we used their official open source implementation: Terminal Bench, SWE-Bench, Scale AI Multi Challenge.

For mathematical and STEM reasoning, we evaluate on \textsc{AIME25} (with and without tools), \textsc{GPQA}~\citep{rein2023gpqa}, \textsc{LiveCodeBench v6}~\citep{jain2024livecodebench},
\textsc{SciCode}~\citep{tian2024scicoderesearchcodingbenchmark},
and \textsc{Humanity’s Last Exam}~\citep{phan2025humanitysexam}.
We additionally include \textsc{MMLU-Pro} to assess general academic and
knowledge-intensive reasoning.

Agentic and tool-augmented capabilities are measured using \textsc{TerminalBench},
\textsc{SWE-Bench} (OpenHands) \citep{jimenez2023swe, wang2025openhandsopenplatformai}, 
\textsc{TauBench V2} (airline, retail, telecom) \citep{barres2025tau}, 
and \textsc{BFCL v4} \citep{patil2025bfcl}, each of which provides verifiable reward signals via
unit tests, database state transitions, or structured schema constraints.

Instruction-following and conversational ability are evaluated with \textsc{IFBench},
\textsc{Scale AI Multi-Challenge}, and \textsc{Arena-Hard-V2} \citep{li2024crowdsourced}. 
These benchmarks probe multi-constraint instructions, preference-aligned chat behavior, and faithfulness to user intent. For \textsc{Arena-Hard-V2}, we follow \citet{yang2025qwen3technicalreport} and use GPT-4.1 as judge.

Long-context performance is assessed with \textsc{RULER-100} at 256k, 512k, and 1M tokens \citep{hsieh2024ruler},
together with \textsc{AA-LCR}, evaluating retrieval, stability, and chain-of-thought
coherence over extreme context lengths. RULER-100 is evaluated with reasoning off, whereas AA-LCR is measured with reasoning on.

For multilingual capability, we report results on \textsc{MMLU-ProX} \citep{xuan2025mmlu} and
\textsc{WMT24++} (en$\rightarrow$xx) \citep{deutsch2025wmt24++}, covering a mix of reasoning and translation
settings across multiple high-resource languages.

For comparison with \textsc{GPT-OSS 20B} and \textsc{Qwen3-30B-A3B-Thinking-2507}, we use the officially reported numbers whenever available; if a benchmark is not reported, we take the value from ArtificialAnalysis (AA)\footnote{\url{https://artificialanalysis.ai/}}; and if neither source provides results, we may compute the scores ourselves using the official evaluation protocol.

Table \ref{tab:nano_v3_comparison} presents a comprehensive performance comparison between the three models. \ourmodel shows strong results, surpassing both GPT-OSS 20B and Qwen3-30B-A3B-Thinking-2507 in all categories. On reasoning benchmarks \ourmodel surpasses the Qwen3 model and is competitive with GPT-OSS, which was the previous best model in these categories. In the agentic, chat, and long context categories \ourmodel significantly outperforms both of the other models, demonstrating the strength of our post-training pipeline.

\section{Quantization}
\label{sec:quantization}

After post-training the model in BF16, we applied Post-Training Quantization (PTQ) using \mbox{ModelOpt}\footnote{\url{https://github.com/NVIDIA/Model-Optimizer}} and Megatron-LM to quantize the model to FP8.

\subsection{Post-Training Quantization Calibration Dataset}
For PTQ calibration, we used a small subset containing 1K samples from post-training reasoning SFT dataset. Using calibration data based on the post-training SFT data yielded slightly better accuracy recovery compared to the \texttt{cnn\_dailymail} dataset.

We also ablated with PTQ using calibration data curated from on-policy generations from the BF16 model, but did not observe any benefit in accuracy recovery compared to the SFT-based calibration dataset. 

\subsection{Selective Post-Training Quantization}
To preserve accuracy while improving efficiency, we used a selective quantization strategy.
We performed quantization sensitivity analysis and explored a set of quantization configurations for mixed-precision models. This study showed that self-attention layers (6 out of 52 layers for \ourmodel) are the most sensitive components, hence we keep them in BF16. Also, the Mamba layers that feed into the self-attention layers were found to be sensitive and are kept in BF16. Overall, keeping the 6 self-attention layers and the 6 Mamba layers in BF16 provided a sweet-spot configuration for accuracy recovery and efficiency trade-off.

The model weights, activations, and KV cache are quantized to FP8. Conv1D within all the Mamba layers are kept in BF16. 

\subsection{Accuracy and Throughput} 
\label{subsec:quantaccuracy}
Table~\ref{tab:nano_v3_quant_accuracy} compares accuracy numbers of \ourmodel FP8 with BF16 on multiple benchmarks. Overall, the FP8 model achieves approximately 99\% median accuracy recovery compared to the BF16 model.

To verify the effectiveness of our selective quantization strategy and to better understand the accuracy–efficiency trade-off, we evaluated several quantization configurations. We conducted an ablation study by applying PTQ to different model components.
Specifically, we examined three factors: attention layer quantization (BF16 or FP8), Mamba layer quantization (FP8 or a mix of BF16 and FP8), and KV cache quantization (BF16 or FP8).

\begin{figure}[h]
    \centering
    \includegraphics[width=\linewidth]{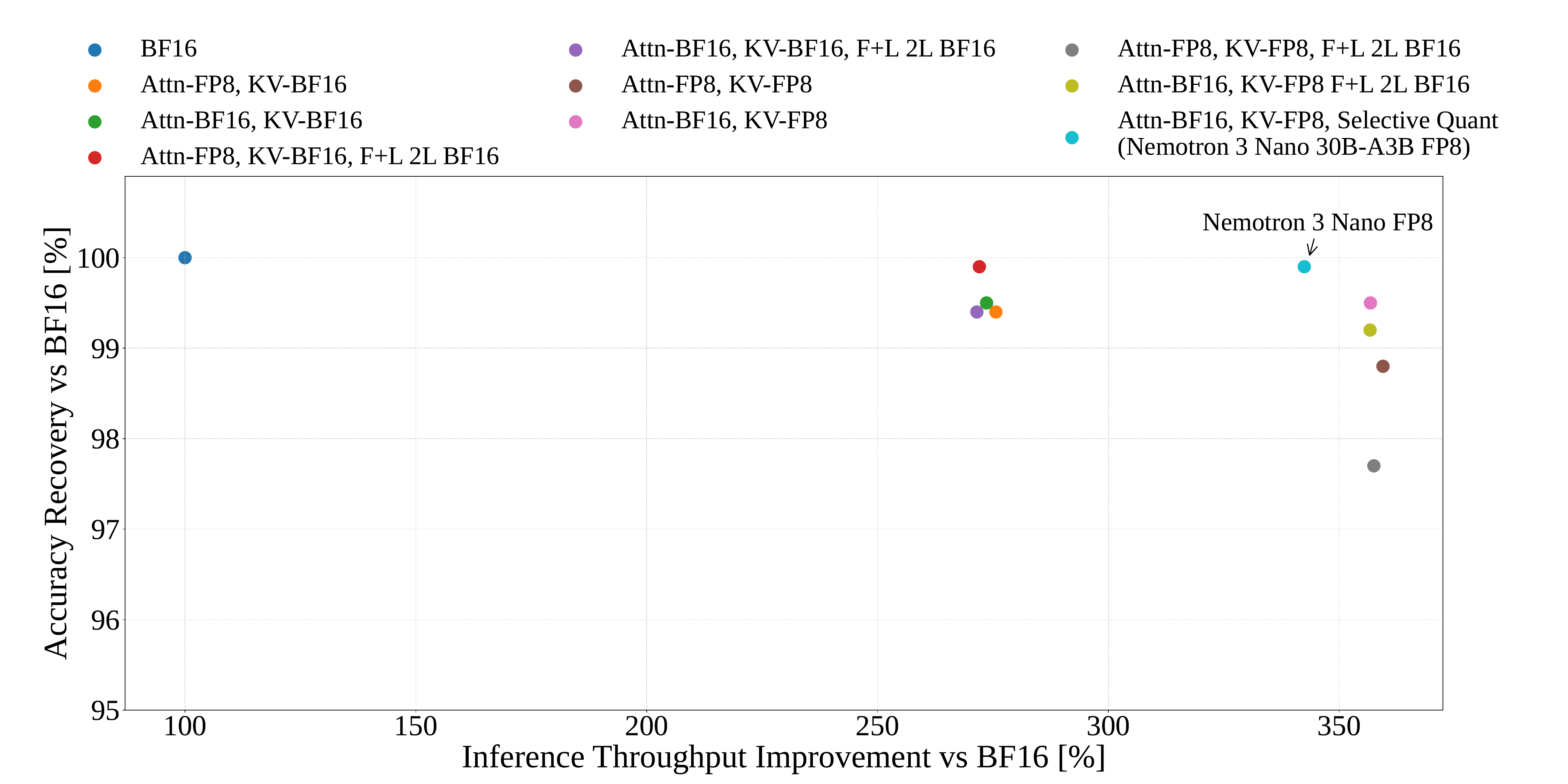}
    \caption{Ablation study of different quantization configurations for accuracy–throughput trade-offs. Accuracy recovery and throughput improvements are computed relative to the \ourmodel{} BF16 checkpoint, with values normalized such that the BF16 baseline is 100\%. Accuracy recovery is defined as the median of the recovery rates across all benchmarks.
    The benchmark was conducted on a single H100 with ISL/OSL=8K/16K. Given that more aggressively quantized models can accommodate larger batch sizes due to lower memory footprint, we used the maximum batch size for each quantization configuration for fair comparisons under the same hardware constraints.}
    \label{fig:fp8_acc_vs_throughput}
\end{figure}

As shown in Figure~\ref{fig:fp8_acc_vs_throughput}, KV cache with FP8 quantization significantly improves throughput by enabling larger batch sizes. While other quantization configurations suffer from accuracy degradation, our selective quantization can retain the accuracy numbers even with KV cache quantization. 
The results confirm that retaining the self-attention layers and their preceding Mamba layers in BF16, while quantizing the remaining layers and the KV cache in FP8, yields a strong accuracy–efficiency trade-off.

\begin{table*}[!htp]
\centering
\small
\setlength{\tabcolsep}{7pt}
\renewcommand{\arraystretch}{1.15}

\begin{tabular}{l|c c }
\toprule
\textbf{Benchmark} & \textbf{N-3-Nano BF16} & \textbf{N-3-Nano FP8} \\
\midrule
\rowcolor{black!5}
\multicolumn{3}{l}{\textbf{General Knowledge}} \\
MMLU-Pro &  78.30 & 78.10 \\

\midrule
\rowcolor{black!5}
\multicolumn{3}{l}{\textbf{Reasoning}} \\
AIME25 (no tools) & 89.06 & 87.71 \\
AIME25 (with tools) & 99.17 & 98.80 \\
GPQA (no tools) &  73.04 & 72.47 \\
GPQA (with tools) & 75.00 & 73.40  \\
LiveCodeBench (v6 2024-08$\leftrightarrow$2025-05) & 68.25 & 67.62 \\
SciCode (subtask) & 33.28 & 31.88 \\
HLE (no tools) & 10.57 & 10.33 \\
HLE (with tools) & 15.48 & 14.27  \\
\midrule
\rowcolor{black!5}
\multicolumn{3}{l}{\textbf{Agentic}} \\
\textbf{TauBench V2} &  &   \\
\quad Airline & 48.00 & 44.79 \\
\quad Retail & 56.91 & 55.59 \\
\quad Telecom & 42.21 & 40.75 \\
\quad Average & 49.04 & 47.04 \\
BFCL v4 & 53.76 & 53.15 \\
\midrule
\rowcolor{black!5}
\multicolumn{3}{l}{\textbf{Chat \& Instruction Following}} \\
IFBench (prompt) & 71.51 & 72.19 \\

\midrule
\rowcolor{black!5}
\multicolumn{3}{l}{\textbf{Long Context}} \\
AA-LCR & 35.85 & 36.06 \\

\midrule
\rowcolor{black!5}
\multicolumn{3}{l}{\textbf{Multilingual}} \\
MMLU-ProX (avg over langs) & 59.50 & 59.63 \\

\bottomrule
\end{tabular}

\caption{Accuracy numbers of \ourmodel{} before/after FP8 quantization.}

\label{tab:nano_v3_quant_accuracy}
\end{table*}


\section{Conclusion}
We present \ourmodel, an open and efficient MoE Hybrid Mamba-Transformer model for agentic reasoning. \ourmodel achieves better or on-par accuracy than competitive models while having up-to 3.3$\times$ higher inference throughput. \ourmodel supports context lengths of up to 1M tokens. We have released the weights for both the base (\ourbasemodel) and final (\ourfinalmodel) models on HuggingFace. Along with the weights, we have also open-sourced the training recipe, data, and code.
\label{sec:conclusion}

\section*{Contributors}

We thank the following people for their invaluable contributions to \ourmodelfull.

\textbf{Pretraining Data.} Abhinav Khattar, Aleksander Ficek, Alisa Liu, Arham Mehta, Asif Ahamed, Ayush Dattagupta, Benedikt Schifferer, Brandon Norick, Branislav Kisacanin, Dan Su, Dane Corneil, Daria Gitman, Dhruv Nathawani, Dima Rekesh, Divyanshu Kakwani, Edgar Minasyan, Eileen Long, Ellie Evans, Eric Tramel, Evelina Bakhturina, Felipe Soares, Feng Chen, Gantavya Bhatt, George Armstrong, Igor Gitman, Ivan Moshkov, Jane Polak Scowcroft, John Kamalu, Johnny Greco, Joseph Jennings, Jupinder Parmar, Kezhi Kong, Markus Kliegl, Maarten Van Segbroeck, Matvei Novikov, Mehrzad Samadi, Miguel Martinez, Mohammad Shoeybi, Mostofa Patwary, Nabin Mulepati, Oleksii Hrinchuk, Rabeeh Karimi Mahabadi, Rima Shahbazyan, Riyad Islam, Roger Waleffe, Rohit Watve, Sadegh Mahdavi, Sanjeev Satheesh, Sean Narentharen, Shrimai Prabhumoye, Shubham Pachori, Shubham Toshniwal, Shuoyang Ding, Somshubra Majumdar, Stephen Ge, Sumeet Kumar Barua, Suseella Panguluri, Syeda Nahida Akter, Vahid Noorozi, Vitaly Kurin, Vitaly Lavrukhin, Wasi Uddin Ahmad, Wei Du, Wei Ping, Yejin Choi, Yev Meyer, Ying Lin, Zihan Liu

\textbf{Architecture.} Abhinav Khattar, Bita Darvish Rouhani, Deepak Narayanan, Ilya Loshchilov, Jatin Mitra, Joey Guman, Mohammad Shoeybi, Mostofa Patwary, Kezhi Kong, Krishna C. Puvvada, Maor Ashkenazi, Nidhi Bhatia, Pavlo Molchanov, Rabeeh Karimi Mahabadi, Rasoul Shafipour, Ritika Borkar, Roger Waleffe, Ryan Prenger, Sanjeev Satheesh, Venmugil Elango, Yonggan Fu

\textbf{Pretraining Software.} Aarti Basant, Ashwath Aithal, Abhinav Khattar, Deepak Narayanan, Duncan Riach, Eric Harper, Hexin Wang, Jared Casper, Jimmy Zhang, Kezhi Kong, Mike Chrzanowski, Nima Tajbakhsh, Pranav Prashant Thombre, Roger Waleffe, Russell J. Hewett, Seonmyeong Bak, Shiqing Fan, Vijay Korthikanti, Xiaowei Ren, Yashaswi Karnati, Zijie Yan

\textbf{Pretraining.} Abhinav Khattar, Brandon Norick, Dan Su, Eric Tramel, Deepak Narayanan, John Kamalu, Joseph Jennings, Jupinder Parmar, Markus Kliegl, Miguel Martinez, Mohammad Shoeybi, Mostofa Patwary, Kezhi Kong, Kevin Shih, Rabeeh Karimi Mahabadi, Roger Waleffe, Ryan Prenger, Shrimai Prabhumoye, Sanjeev Satheesh, Syeda Nahida Akter, Ying Lin 

\textbf{Long Context.} Boris Ginsburg, Cheng-Ping Hsieh, Dan Su, Dima Rekesh, Faisal Ladhak, Fei Jia, John Kamalu, Kezhi Kong, Krishna C. Puvvada, Markus Kliegl, Mostofa Patwary, Roger Waleffe, Samuel Kriman, Sanjeev Satheesh, Shantanu Acharya, Simeng Sun, Ushnish De

\textbf{Posttraining Software.} Adi Renduchintala, Alexander Bukharin, Ali Taghibakhshi, Banghua Zhu, Brian Yu, Duncan Riach, Frankie Siino, Gerald Shen, Jiaqi Zeng, Kezhi Kong, Li Ding, Luis Vega, Maanu Grover, Marc Romeijn, Parth Chadha, Peter Jin, Soumye Singhal, Terry Kong, Tugrul Konuk, Yi-Fu Wu, Yubo Gao

\textbf{Posttraining.} Abhibha Gupta, Adi Renduchintala, Akanksha Shukla, Aleksander Ficek, Alexander Bukharin, Ameya Sunil Mahabaleshwarkar, Banghua Zhu, Besmira Nushi, Branislav Kisacanin, Cheng-Ping Hsieh, Charles Wang, Damon Mosk-Aoyama, Daria Gitman, Dhruv Nathawani, Dima Rekesh, Edgar Minasyan, Edward Lin, Evelina Bakhturina, Fei Jia, Felipe Soares, Feng Chen, George Armstrong, Grigor Nalbandyan, Haifeng Qian, Hayley Ross, Igor Gitman, Ivan Moshkov, Jeffrey Glick, Jiaqi Zeng, Jian Zhang, Jie Lou, Julien Veron Vialard, Junkeun Yi, Katherine Luna, Khushi Bhardwaj, Krishna C. Puvvada, Luis Vega, Makesh Narsimhan Sreedhar, Matvei Novikov, Mehrzad Samadi, Mengru Wang, Michael Evans, Nikolai Ludwig, Oleksii Hrinchuk, Oleksii Kuchaiev, Olivier Delalleau, Ouye Xie, Peter Jin, Pritam Gundecha, Prasoon Varshney, Rima Shahbazyan, Ritu Gala, Sadegh Mahdavi, Sahil Modi, Sanjay Kariyappa, Sean Narenthiran, Shantanu Acharya, Shubham Toshniwal, Shuoyang Ding, Somshubra Majumdar, Soumye Singhal, Stephen Ge, Sugam Dipak Devare, Suseella Panguluri, Tugrul Konuk, Vahid Noroozi, Venkat Srinivasan, Vitaly Lavrukhin, Wasi Uddin Ahmad, Wei Du, Yev Meyer, Yian Zhang, Yoshi Suhara

\textbf{Evaluation, Safety and Release.} Aaron Grattafiori, Barnaby Simkin, Besmira Nushi, Bilal Kartal, Christopher Parisien, Daniel Rohrer, David Mosallanezhad, Eileen Peters Long, Erick Galinkin, Fay Wang, Ferenc Galko, Gorkem Batmaz, Jane Polak Scowcroft, Katherine Luna, Khushi Bhardwaj, Leon Derczynski, Michael Boone, Michael Evans, Piotr Januszewski, Rich Harang, Rishabh Garg, Riyad Islam, Sanjay Kariyappa, Sanjeev Satheesh, Shaona Ghosh, Wojciech Prazuch, Yoshi Subara, Zhen Dong, Zijia Chen

\textbf{Infrastructure.} Aaron Blakeman, Anubhav Mandarwal, Alex Kondratenko, Aleksandr Shaposhnikov, Ashwin Poojary, Brandon Soubasis, Collin Neale, Dong Ahn, Evan Briones, Gargi Prasad, Harsh Sharma, Herman Sahota, Himanshu Soni, Jining Huang, Kumar Anik, Maer Rodrigues de Melo, Nikhil Jukar, Pasha Shamis, Rick Izzo, Ruoxi Zhang, Satish Pasumarthi, Sergey Kashirsky, Shelby Thomas, Stefania Alborghetti

\textbf{Quantization.} Aditya Vavre, Akhiad Bercovich, Ameya Sunil Mahabaleshwarkar, Amnon Geifman, Asma Kuriparambil Thekkumpate, Ben Lanir, Bilal Kartal, Chenhan Yu, Daniel Afrimi, Darko Stosic, Dusan Stosic, Ganesh Ajjanagadde, Huizi Mao, Ido Shahaf, Jenny Chen, Kai Xu, Nave Assaf, Omer Ullman Argov, Ran Zilberstein, Sharath Turuvekere Sreenivas, Sweta Priyadarshi, Tijmen Blankevoort, Tomer Asida, Yoshi Suhara,  Zach Moshe, Zijia Chen

\textbf{Inference.} Amir Klein, Amit Zuker, Chenghao Zhang, Daniel Afrimi, Daniel Serebrenik, Gal Hubara Agam, Helen Ngo, Joyjit Daw, Kan Zhu, Keshav Santhanam, Lawrence McAfee, Lucas Liebenwein, Luis Vega, Nave Assaf, Neta Zmora, Netanel Haber, Omer Ullman Argov, Peter Dykas, Pranav Prashant Thombre, Ran Zilberstein, Roi Koren, Shahar Mor, Shanmugam Ramasamy, Siddharth Singh, Suyog Gupta, Teodor-Dumitru Ene, Tomer Asida, Tomer Bar Natan, Vijay Korthikanti, Wanli Jiang, William Zhang, Yashaswi Karnati

\textbf{Deployment.} Alexandre Milesi, Anahita Bhiwandiwalla, Huy C Nguyen, Huy Q Nguyen, Izzy Putterman, Manoj Kilaru, Maryam Moosaei, Pawel Morkisz, Tan Bui, Thanh Do

\textbf{Legal and Compliance.} Barnaby Simkin, Chantal Hwang, Chetan Mungekar, Dina Yared, Hiren Upadhyay, Iain Cunningham, Katherine Cheung, Laya Sleiman, Meredith Price, Michael Boone, Nikki Pope, Saori Kaji

\textbf{Marketing.} Amelia Barton, Chintan Patel, Erik Pounds, Mark Cai, Natalie Hereth, Nicola Sessions, Nirmal Juluru, Shreya Gopal, Will Jennings

\textbf{Project Management.} Amy Shen, Ann Guan, Bardiya Sadeghi, Daria Levy, Elena Lantz, Elliott Ning, Krzysztof Pawelec, Melissa Corpuz, Negar Habibi, Pinky Xu, Qing Miao, Ryan Timbrook, Seth Poulos, Smita Ithape, Twinkle Vashishth

\textbf{Product.}  Chris Alexiuk, Ellie Evans, Jane Polak Scowcroft, Jesse Oliver, Joey Conway, Tom Balough, Udi Karpas, Wenfei Zhou

\textbf{Leadership.} Andrew Tao, Bita Darvish Rouhani, Boris Ginsburg, Bryan Catanzaro, Carlo del Mundo, Eileen Long, Eric Chung, Jane Polak Scowcroft, Jan Kautz, Jian Zhang, Joey Conway, Jonathan Cohen, Kari Briski, Mohammad Shoeybi, Mostofa Patwary, Oleksii Kuchaiev, Oluwatobi Olabiyi, Pavlo Molchanov, Ran El-Yaniv, Ran Zilberstein, Yonatan Geifman, Yejin Choi

\newpage

\bibliography{references}
\bibliographystyle{references}

\appendix

\section{Base model evaluations}
\label{appendix:v0_base_model_evals}

For completeness, Table~\ref{tab:base-model-comparison-reproducible} presents the evaluation results for the base model checkpoint used to initialize the alignment process (referred to as the pre-alignment base). During development, we identified limitations in this model's performance on few key benchmarks, which motivated the training of the improved base model intended for release (as evaluated in Table~\ref{tab:model-comparison}).

Unlike the pre-alignment base, which trailed Qwen3, the improved checkpoint surpasses Qwen3 in the average performance on Code and General Knowledge tasks. The lead in Math tasks has also significantly widened compared to the pre-alignment base. In Multilingual benchmarks, while Qwen3 retains a lead on the MMLU Global Lite task, the improved checkpoint has surpassed Qwen3 on the MGSM task. The only significant regression is in Long Context, where the improved checkpoint shows a slight performance drop compared to the pre-alignment base, though it still maintains a commanding margin over Qwen3.

\begin{table}[!hbt]
\centering
\small
\setlength{\tabcolsep}{7pt}
\renewcommand{\arraystretch}{1.15}

\begin{tabular}{l|cc}
\toprule
\textbf{Task} &
\textbf{Qwen3} &
\textbf{N-3-Nano-Pre-Align} \\
\midrule

\rowcolor{black!5}
\multicolumn{3}{l}{\textbf{General Knowledge}} \\
MMLU (5-shot, acc) & \textbf{81.07} & 78.44 \\
MMLU-Pro (5-shot, CoT EM) & \textbf{61.71} & 61.39 \\
AGIEval-En (3/5-shot, CoT acc) & 63.12 & \textbf{65.62} \\
\midrule
\rowcolor{black!5}
\multicolumn{3}{l}{\textbf{Code}} \\
HumanEval (0-shot) & \textbf{70.73} & 69.51 \\
MBPP-Sanitized (3-shot) & \textbf{73.15} & 71.21 \\
\midrule
\rowcolor{black!5}

\multicolumn{3}{l}{\textbf{Math}} \\
GSM8K (8-shot, acc) & \textbf{89.01} & 87.04 \\
MATH (4-shot, acc) & 61.14 & \textbf{80.80} \\
MATH-500 (4-shot, avg@32) & 55.08 & \textbf{72.79} \\
\midrule
\rowcolor{black!5}

\multicolumn{3}{l}{\textbf{Commonsense Understanding}} \\
ARC-Challenge (25-shot, acc\_norm) & \textbf{94.45} & 91.81 \\
HellaSwag (10-shot, acc\_norm) & 83.14 & \textbf{86.08} \\
OpenBookQA (0-shot, acc\_norm) & 44.80 & \textbf{46.60} \\
PIQA (0-shot, acc\_norm) & 81.01 & \textbf{83.68} \\
WinoGrande (5-shot, acc) & 78.22 & \textbf{79.08} \\
\midrule
\rowcolor{black!5}

\multicolumn{3}{l}{\textbf{Reading Comprehension}} \\
RACE (0-shot, acc) & \textbf{90.05} & 87.56 \\
\midrule
\rowcolor{black!5}

\multicolumn{3}{l}{\textbf{Multilingual}} \\
MMLU Global Lite (5-shot, avg acc) & \textbf{76.84} & 75.69 \\
MGSM (8-shot, avg acc) & \textbf{82.53} & 78.93 \\
\midrule
\rowcolor{black!5}

\multicolumn{3}{l}{\textbf{Long Context}} \\
RULER (64K, 0-shot, acc) & 63.55 & \textbf{88.94} \\
RULER (128K, 0-shot, acc) & 60.69 & \textbf{86.78} \\
RULER (256K, 0-shot, acc) & - & \textbf{79.15} \\
\bottomrule
\end{tabular}
\caption[Comparison of Qwen3 vs N-Nano-3]{
    Comparison of \textbf{Qwen3-30B-A3B-Base} and the \textbf{\ourmodel} pre-alignment base checkpoint (the specific checkpoint used to initialize the alignment pipeline). Best results between these two are marked in bold. 
}
\label{tab:base-model-comparison-reproducible}
\end{table}

\section{MMLU-redux evaluation}
\label{appendix:mmlu_redux_evaluation}

We developed the following two variants of MMLU-redux:

(1) MMLU-redux CoT. We created this variant due to the observation that many STEM questions intrinsically require step-by-step reasoning for successful resolution, which is not adequately captured by the original multiple-choice, no chain-of-thought format. The model might arrive at some answers through guessing or memorization. Therefore, we created five exemplars per subject, each accompanied by a detailed step-by-step solution. This allows us to evaluate models using a 5-shot chain-of-thought setting.

(2) MMLU-redux Tweak.
As MMLU's widespread use increases the risk of overfitting and benchmark saturation from extensive tuning, we introduced this variant to more rigorously evaluate model performance on similar yet new examples that closely match the original in difficulty, style, structure, and format.
We modified the original test examples using Qwen3-235B-A22B-Thinking-2507 to assess the same underlying concepts, ideas, and skills while altering specific details such as numerical values and equations.

The evaluation results are presented in Table~\ref{table:ablation_mmlu_redux_variants}.
Overall, enabling CoT reasoning yields a substantial accuracy boost, especially on STEM subjects.
Our model demonstrates a larger gain from CoT compared to Qwen (an average improvement of $+5.27$ versus $+0.79$, respectively).
In addition, we observe a significant increase in the Professional Accounting task under the Other category, with an improvement from $64.00$ to $77.00$ ($+13.00$), as this task also relies heavily on calculation skills.

On MMLU‑redux Tweak, both models achieve noticeable gains across non‑STEM categories, likely because many non‑STEM questions assess domain knowledge, and the tweaked questions were generated using Qwen3‑235B‑A22B‑Thinking‑2507, whose knowledge may align more closely with the evaluated models.
We observe a divergent trend on STEM: Qwen's accuracy decreases marginally ($-0.83$), while our model's score increases by $5.31$.

\begin{table*}[!hbt]
\small
\centering
\setlength{\tabcolsep}{6pt}
\newcolumntype{F}{>{\centering\arraybackslash}p{1.2cm}} 
\newcolumntype{C}{>{\centering\arraybackslash}p{2.3cm}} 
\begin{tabular}{l|F|F|C|C|C|C}
\toprule
& \multicolumn{2}{c|}{\textbf{MMLU-redux}} & \multicolumn{2}{c|}{\textbf{MMLU-redux CoT}} & \multicolumn{2}{c}{\textbf{MMLU-redux Tweak}} \\
 & \multicolumn{1}{c|}{Qwen} & \multicolumn{1}{c|}{Ours} & \multicolumn{1}{c|}{Qwen} & \multicolumn{1}{c|}{Ours} & \multicolumn{1}{c|}{Qwen} & \multicolumn{1}{c}{Ours} \\
\midrule
STEM & 81.05 & 74.42 & 84.05 ($+3.00$) & 87.26 ($+12.84$) & 80.22 ($-0.83$) & 79.26 ($+4.84$) \\
Humanities & 82.31 & 80.46 & 83.16 ($+0.85$) & 81.23 ($+0.77$) & 85.04 ($+2.73$) & 84.04 ($+3.58$) \\
Social Sciences & 86.83 & 84.42 & 85.92 ($-0.91$) & 85.50 ($+1.08$) & 89.36 ($+2.53$) & 89.70 ($+5.28$) \\
Other & 80.23 & 77.85 & 79.85 ($-0.38$) & 80.38 ($+2.53$) & 82.43 ($+2.20$) & 84.00 ($+6.15$) \\
All & 82.37 & 78.68 & 83.16 ($+0.79$) & 83.95 ($+5.27$) & 83.76 ($+1.39$) & 83.64 ($+4.96$) \\
\bottomrule
\end{tabular}
\caption{Evaluation results on MMLU-redux and two variants. ``Qwen'' refers to the Qwen3-30B-A3B-Base model. ``Ours'' denotes our base model checkpoint used in the ablation study, which was trained on a data blend that differs slightly from the one used for our final model as the ablation study was conducted alongside training.}
\label{table:ablation_mmlu_redux_variants}
\end{table*}

\section{DPO for Reducing Tool Hallucination}

Reducing hallucinated tool usage is one of the key objectives of our alignment experiments. 
Although our released model does not rely on DPO, because reinforcement learning (RL) already achieved comparable performance, we nevertheless explored DPO as an additional technique due to its simplicity and minimal computational overhead. As shown later, even a very small amount of DPO training yields meaningful reductions in hallucinated tool calls and improves reasoning stability. To support this analysis, we first define what constitutes hallucinated tool usage in our evaluation.

\noindent\textbf{Definition of Tool Hallucination and Hallucination Rate.}  
We define \textbf{tool hallucination} as any instance in which the model attempts to invoke a tool despite no tools being declared in the system message. Under the \textit{No-Tools} and \textit{Hallucination-Penalty} settings, the model is expected to rely entirely on internal reasoning; therefore, any output containing a tool call, such as a Python execution request, a search invocation, or any tool-specific API format, is treated as a hallucination.

The \textbf{tool hallucination rate} is the proportion of evaluation samples in which such unintended tool calls occur. A higher rate indicates inappropriate tool triggering, whereas a near-zero rate reflects strong calibration and reliable adherence to environment constraints.

\noindent\textbf{DPO Data Construction.}  
To study how DPO affects tool-use calibration and reasoning performance, we constructed a DPO dataset using 2{,}000 reasoning tasks: 1{,}000 mathematics problems and 1{,}000 STEM multi-choice questions. For each problem, the model generated 32 on-policy solutions, providing a diverse set of candidate behaviors. These raw generations were then processed through our DPO data-construction pipeline, assigning preference labels according to correctness and tool-usage conditions, which produced approximately 50k preference samples in total. We later found that the model’s improvements persisted even when using substantially smaller datasets; in fact, training with as few as 10k preference samples (or even fewer) yielded similar benefits. This further underscores the low computational cost and high sample efficiency of DPO in our setting. To study tool-use alignment, we organized the data into three categories: (1) No-Tools, where the system message does not expose tools and correctness alone determines preference labels; (2) With-Tools, where tools are available and labels depend only on the correctness of the final answer; and (3) Hallucination-Penalty, where tools are not declared and any hallucinated tool invocation is labeled as a negative preference. This structure allows us to jointly evaluate pure reasoning ability, tool-assisted reasoning, and calibration of tool usage, while providing a rich set of preference signals derived from diverse on-policy model behaviors.

\noindent\textbf{Training Setup.}  
For our DPO experiments, we used a lightweight training configuration designed to minimally perturb the model after SFT while still providing a meaningful preference-learning signal. Specifically, we trained with a learning rate of 3e-6, a batch size of 128, and 50 training steps. We set the SFT loss coefficient to 0.2, the preference (DPO) loss coefficient to 1.0, and the KL loss coefficient to 0.05. This setup emphasizes preference learning while retaining a small supervised loss to stabilize outputs and a modest KL penalty to prevent excessive deviation from the base model.
This configuration emphasizes preference learning while retaining a small supervised loss to stabilize outputs and a modest KL penalty to prevent excessive deviation from the base model.

\noindent\textbf{Results.}  
Table \ref{table:dpo-results} shows the impact of applying a small amount of DPO training on both reasoning accuracy and hallucinated tool usage. Despite using only 50 training steps with a modest learning rate, we observe consistent improvements across all evaluated benchmarks.

For AIME25, accuracy increases from 80.88\% to 84.58\%, indicating that DPO not only suppresses undesirable tool-related behaviors but also enhances overall solution quality. Notably, the hallucination rate, which is already low in this setting, is reduced from 1.25\% to 0\%, fully eliminating spurious tool invocation.

On GPQA, which is more challenging and shows higher baseline hallucination, DPO again yields substantial gains. Accuracy improves from 65.15\% to 69.19\%, and the hallucination rate drops dramatically from 8.33\% to just 0.7\%. This confirms that preference-based fine-tuning is particularly effective in settings where the model is prone to uncertainty or over-triggering tool calls.

Overall, the results demonstrate that even minimal DPO training can meaningfully reduce hallucinated tool usage while simultaneously improving reasoning accuracy. This suggests that DPO provides a valuable complementary signal to RL-based alignment, strengthening both model reliability and calibration with negligible computational cost.

\begin{table*}[!hbt]
\small
\centering
\setlength{\tabcolsep}{6pt}
\newcolumntype{F}{>{\centering\arraybackslash}p{1.2cm}}
\newcolumntype{C}{>{\centering\arraybackslash}p{2.3cm}}
\begin{tabular}{l|F|F|C|C}
\toprule
& \multicolumn{2}{c|}{\textbf{Accuracy}} & \multicolumn{2}{c}{\textbf{Hallucination Rate}} \\
 & \multicolumn{1}{c|}{Before DPO} & \multicolumn{1}{c|}{After DPO} 
 & \multicolumn{1}{c|}{Before DPO} & \multicolumn{1}{c}{After DPO} \\
\midrule
AIME25 (no tools) & 80.88 & 84.58 & 1.25\% & 0\% \\
GPQA (no tools)  & 65.15 & 69.19 & 8.33\% & 0.7\% \\
\bottomrule
\end{tabular}
\caption{Evaluation results on DPO experiments.}
\label{table:dpo-results}
\end{table*}

\section{Safety Preference Data}
\label{sec:safety_pref_data}
For the RLHF stage, reward model training data comprises of the same underlying datasets used in the SFT safety subset, leading to a similar distribution for the starting seed prompts. Response generation is more nuanced, to handle over-refusals and harmful engagements as the rejected responses. 
\begin{itemize}
\vspace{-1em}
    \item For \textbf{harmful prompts}, chosen responses are generated with a similar strategy as the SFT responses. The rejected responses are unsafe model outputs, generated via two methods: (i) applying jailbreak templates to produce harmful completions, and (ii) directly prompting the model and using a content safety moderation classifier to detect cases of harmful outputs.
    \item For \textbf{safe prompts}, chosen responses are generated by passing the safe prompt as-is to the underlying model, and using a content safety moderation classifier to ensure safe responses. The rejected responses are generated by applying refusal prompt templates, resulting in over-refusals.
\end{itemize} 
\vspace{-1em}
The resulting response pairs are thus annotated using a preference-based scheme: for harmful prompts, <safe, unsafe> completions are labeled as the <chosen, rejected> pairs. For safe prompts, <safe, over-refusal> completions are annotated similarly as <chosen, rejected> pairs. This approach supports training reward models for both robust safety alignment and mitigating over-refusal behaviors.

To ensure diversity, we generate the chosen and rejected response pairs for each prompt using five (5) different open-source models, followed by applying necessary filters to keep only safe (for chosen) and unsafe or over-refusal responses (for rejected) to build a list of candidate chosen and rejected responses. Finally, one chosen and rejected response pair per prompt is chosen randomly from the candidates.

\section{Prompt Sensitivity Analysis}
\label{sec:prompt_sens}

\begin{table*}[!htp]
\centering
\small
\setlength{\tabcolsep}{7pt}
\renewcommand{\arraystretch}{1.15}

\begin{tabular}{l|c c c }
\toprule
\textbf{Benchmark} & \textbf{N-3-Nano} & \textbf{Qwen3} & \textbf{GPT-OSS} \\
\midrule

GPQA (no tools) & 0.42 & 0.59 & 1.91 \\
MMLU-Pro & 0.41 & 0.31 & 1.46 \\
Comp-Math-24-25 (no tools) & 0.77 & 0.51 & 1.14 \\
LiveCodeBench (v6 2024-08$\leftrightarrow$2025-05) & 0.83 & 1.05 & 1.02\\
\bottomrule
\end{tabular}

\caption{
Prompt sensitivity for \ourmodel, \text{Qwen3-30B-A3B-Thinking-2507} and \text{GPT-OSS 20B} (lower is better). (Comp-Math-24-25 contains AIME24, AIME25, HMMT 2024 Feb., Nov. and 2025 Feb. datasets).
}
\label{tab:prompt_sensitivty}
\end{table*}

LLM predictions can be sensitive to minor changes to the input~\citep{nalbandyan2025score}. Even simple, non-adversarial edits (e.g., changes in prompt wording, answer formatting instructions, or problem placement relative to the prompt) can shift the model’s outputs enough to change individual predictions and, in aggregate, benchmark accuracy. To reduce the risk of over- or under-estimating accuracy due to a single prompt choice, we evaluate models using multiple prompts. This better reflects model stability under routine, realistic prompt variations.

To measure prompt sensitivity, we construct a set of prompts for each dataset varying in wording, instruction granularity (minimal vs. detailed), problem placement (before, middle, or after the prompt), and answer formatting. For each prompt, we compute mean accuracy across eight seeds, and we use the standard deviation of prompt averages as the prompt sensitivity metric. Prompt sensitivity results are presented in Table~\ref{tab:prompt_sensitivty}. With sensitivity scores below 1 across all datasets, \ourmodel shows strong stability and robustness to changes in the prompt.




\end{document}